\documentclass[11pt]{article}

\usepackage[final]{acl}

\usepackage{times}
\usepackage{latexsym}

\usepackage[T1]{fontenc}

\usepackage[utf8]{inputenc}

\usepackage{microtype}

\usepackage{inconsolata}

\usepackage{graphicx}

\title{Learning to Control Summaries with Score Ranking}

\author{Hongye Liu \\
  Duke University \\
  \texttt{hongye.liu@duke.edu} \\\And
  Liang Ding \\
  University of Sydney \\
  \texttt{liangding.liam@gmail.com} \\\And
  Ricardo Henao \\
  Duke University \\
  \texttt{ricardo.henao@duke.edu} \\
  }

\usepackage{hyperref}        
\usepackage{url}            
\usepackage{booktabs}        
\usepackage{amsfonts}        
\usepackage{nicefrac}        
\usepackage{xcolor}     

\usepackage{amsmath}
\usepackage{float}    

\usepackage{multirow}
\usepackage{booktabs}
\usepackage{arydshln}  
\usepackage{wrapfig}

\PassOptionsToPackage{table,xcdraw}{xcolor}

\usepackage{enumitem}
\setlist[itemize,1]{leftmargin=1em}

\setcitestyle{round}

\begin{document}
\maketitle
\begin{abstract}
Recent advances in summarization research focus on improving summary quality across multiple criteria, such as completeness, conciseness, and faithfulness, by jointly optimizing these dimensions.
However, these efforts largely overlook the challenge of controlling summary generation with respect to individual criteria, especially in the presence of their inherent trade-offs.
For example, enhancing conciseness can compromise completeness, and {\em vice versa}.
In this work, we address this gap by proposing a loss function that aligns model outputs with fine-grained, model-based evaluation scores ({\em e.g.}, from FineSurE), enabling both improvement in summary quality and dimension-specific control.
Our approach\footnote{The source code and model are available, respectively, at \href{https://github.com/Hyfred/Control-Summaries-with-Ranking}{Control-Summaries-with-Ranking}
and \href{https://huggingface.co/hongyeeliu/Control_Summaries_LLaMA}{Control-Summaries-LLaMA}.} improves the overall quality of the summaries while maintaining the ability to selectively prioritize one criterion over others.
Experiments on three pretrained models (LLaMA, Qwen, and Mistral) demonstrate that our method achieves performance comparable to state-of-the-art summarizers, while uniquely offering strong controllability over individual quality dimensions. 
\end{abstract}

\section{Introduction}
Automatic text summarization~\citep{hovy2005text,tas2007survey} aims to condense long documents into concise text descriptions that preserve the most important information from the original text, and is widely used in real-world applications~\citep{ji2026strideedstrategygroundedstepwisereasoning, xu2026rcbsf}.
Despite huge progress in neural summarization~\citep{cheng2016neural,kryscinski2019neural}, most systems are trained to optimize surrogate objectives, such as model likelihood or $n$-gram overlap metrics, which correlate poorly with human judgment of semantic fidelity, factual consistency, and content balance~\citep{maynez2020faithfulness,pagnoni2021understanding}.
Recent model-based evaluators like FineSurE~\citep{song2024finesure} and UniSumEval~\citep{lee2024unisumeval} instead extract and align {\em atomic semantic units} or \emph{keyfacts} between the source and summary, delivering fine-grained {\em completeness} and {\em conciseness} scores that correlate better with human assessments~\citep{laban2023summedits}.

\begin{figure}[t]
    \centering
    \includegraphics[width=\columnwidth]{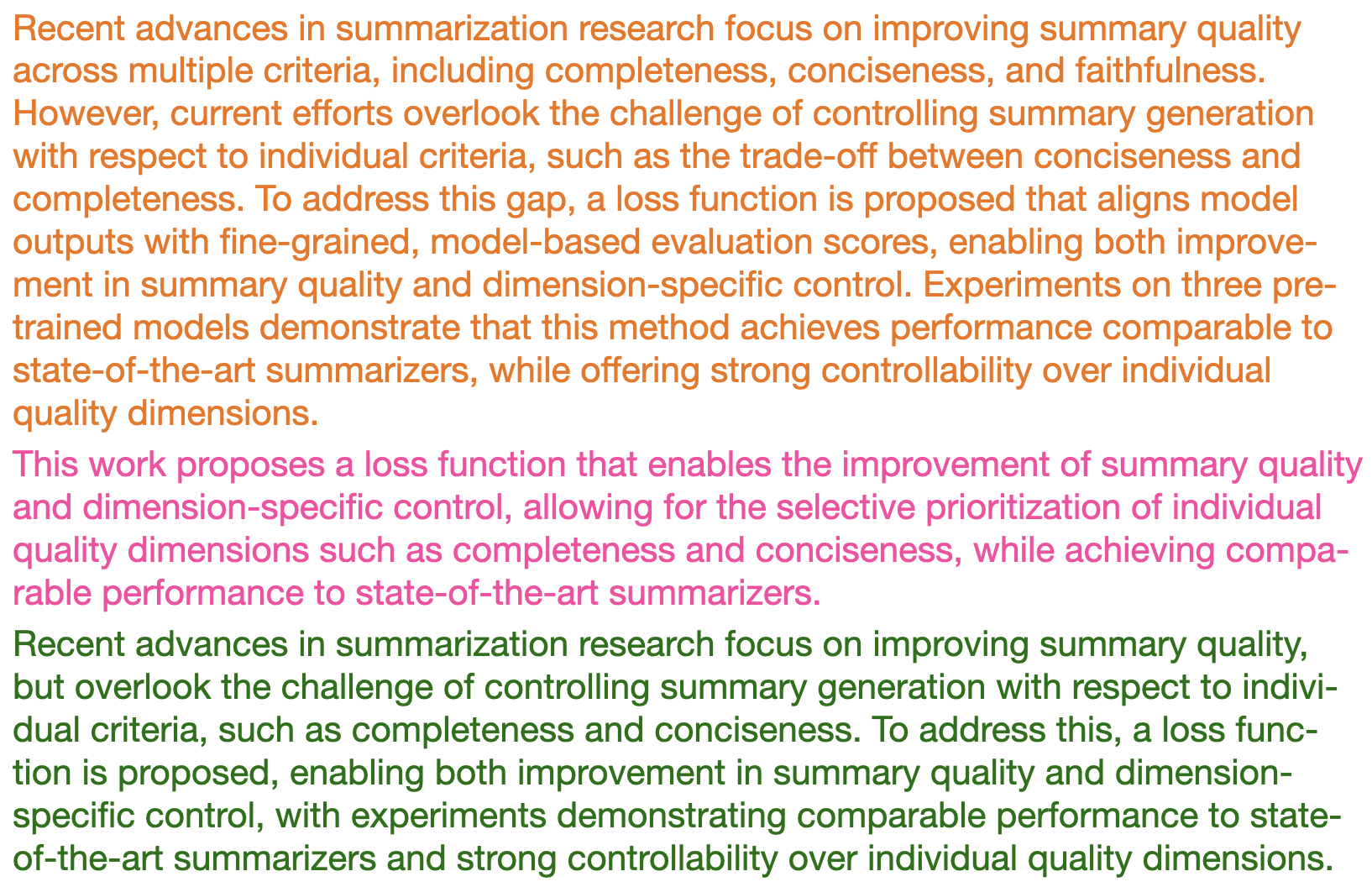}
    \vspace{-6mm}
    \caption{\small 
    Summaries of the abstract prioritizing completeness (\textcolor[HTML]{ff7f0e}{Com$_\uparrow$}), conciseness (\textcolor[HTML]{E22B98}{Con$_\uparrow$}) and balance (\textcolor[HTML]{009901}{Bal}).
    }
    \label{fig:summary_example}
    \vspace{-6mm}
\end{figure}

On the optimization side, reinforcement learning (RL) methods, {\em e.g.}, Proximal Policy Optimization (PPO)~\citep{schulman2017proximal} and Direct Preference Optimization (DPO)~\citep{rafailov2024direct}, have been applied to directly optimize non-differentiable quality metrics; however, they suffer from high variance, instability, or prohibitive computational costs~\citep{ahmadian2024back,liu2024length,yan2024reward}.
More scalable ranking-based objectives, such as those based on contrastive loss \citep{liu2022brio} and margin ranking, encourage higher-quality candidates to outrank inferior ones~\citep{liu2023learning,chern2023improving}.
However, they typically optimize a single aggregated score and do not offer mechanisms to steer summaries along distinct dimensions such as summary {\em completeness} or {\em conciseness}.
In fact, studies on the ``alignment tax'' have shown that improving one dimension often degrades another~\citep{noukhovitch2023language,guo2024controllable}.
Details of the related work are provided in Appendix~\ref{app:related}.

In this work, we propose a framework for \emph{optimizing} neural summarization with explicit control over multiple quality dimensions, with a particular focus on adjusting the trade-off between summary completeness and conciseness. We empirically demonstrate such a trade-off in Appendix~\ref{appendix:tradeoff_verification}.
The key ideas we pursue are:
$i$) {\em fine-grained ranking} by combining a margin-ranking loss to align the model's log-likelihood with model-based quality scores and a maximum-scoring loss to directly push the top candidate toward higher overall quality;
$ii$) {\em control-oriented loss} that, given a prompt indicating the desire for a more complete (Com$_\uparrow$), concise (Con$_\uparrow$) or balanced (Bal) summary, adjusts the ratio of completeness to conciseness scores to meet the specified intent; and
$iii$) {\em unified training objective} that integrates margin ranking, maximum scoring, and control losses, allowing one model to flexibly generate summaries optimized for different trade-offs.
Figure~\ref{fig:summary_example} shows summary examples from our model using the abstract as a source document. 

Technically, we implement our approach by fine-tuning LoRA adapters~\citep{hu2022lora} on three open source backbone models, namely LLaMA~\citep{touvron2023llama, touvron2023llama2, dubey2024llama}, Qwen~\citep{yang2024qwen2} and Mistral~\citep{jiang2023mistral7b}, and evaluate on both {\em in-domain} (WikiHow~\citep{koupaee2018wikihow}, CNN/DM~\citep{nallapati2016abstractive}, and DialogSum~\citep{chen2021dialogsum}), and {\em out-of-domain} tasks (OpoSum~\citep{angelidis2018summarizing}, MeQSum~\citep{abacha2019summarization}).
Experiments show that our method:
$i$) significantly improves the Spearman rank correlation between model likelihood and model-based scores;
$ii$) improves the (harmonic) mean of completeness and conciseness for (test) summaries; and
$iii$) delivers strong controllability, producing summaries with high fidelity to the specified dimension while maintaining faithfulness (consistency wrt source). 
Our contributions can be summarized as follows.
\begin{itemize}[topsep=-1mm,itemsep=-1mm,leftmargin=3mm]
    \item We introduce a joint ranking and scoring framework that aligns generation likelihoods with fine-grained model-based evaluation metrics, achieving superior ranking performance compared to existing baselines.
    \item We propose a control-oriented loss that steers summary generation toward completeness, conciseness, or balance based on simple prompts.
    \item We demonstrate the generality of our approach by fine-tuning multiple language models and evaluating across diverse domains, showing consistent gains in overall quality and controllability.
\end{itemize}

\section{Methodology}\label{section:method}
{\bf Problem Definition}
Our goal is to generate a summary \( Y = (y_1, \ldots, y_N) \) from a source document \( X \) by modeling the conditional likelihood \( p_\theta(Y | X, Z) \), where \( Z \) is a prompt that controls the summarization process.
Following the standard autoregressive modeling framework, the generation process is formulated as follows:
\begin{equation}
p_\theta(Y | X, Z) = \textstyle{\prod}_{i=1}^{N} p_\theta(y_i | y_{<i}, X, Z) \,,
\label{eq:autoregressive}
\end{equation}
where \( y_i \) is the \( i \)-th generated token, \( y_{<i} = (y_1, \ldots, y_{i-1}) \) denotes all previously generated tokens and $N$ is the summary length.
When \( i = 1 \), \( y_{<1} \) is the empty sequence, and the first token is generated conditioned only on \( (X,Z) \).
We seek to generate an optimal summary $\tilde{Y}$ by maximum likelihood (ML), {\em i.e.}, $\tilde{Y} = \arg\max_Y \ p_\theta(Y | X, Z)$.
However, finding the exact ML solution (summary) is intractable due to the exponential size of the output space.
In practice, we approximate this by generating a set of candidate summaries using nucleus sampling~\citep{holtzman2019curious}.

To generate {\em high-quality} summaries, it is desirable to align \(\tilde{Y}\) with a chosen quality score \( S_{D}(\tilde{Y})= S(\tilde{Y} | X, D) \) that evaluates the summary \(\tilde{Y}\) given the document \( X \) for some evaluation criterion \( D \).
In this work, we focus on summary {\em completeness} and {\em conciseness}, as defined in~\citet{song2024finesure}, however, there are several others that can be considered~\citep{lloret2018challenging}. 
Moreover, for generality purposes, we assume that \( S(\cdot) \) is provided by a {\em black-box model}, {\em i.e.}, a learned scorer that assigns quality scores but does not expose (propagate) gradients.
This assumption promotes scalability and modularity, since the scorer can be replaced or improved independently of the generation model and, in principle, can be applied to any generator without retraining.
Previous work has also explored differentiable scoring functions, such as learned reward models for RL~\citep{kaelbling1996reinforcement}, but these approaches require gradient access and often suffer from stability issues during optimization~\citep{schulman2017proximal}.
In the remainder of this work, \( S(\cdot) \) is a model-based score.

Our objective is to model \( p_\theta \) such that the generated summary \( \tilde{Y} \) achieves a high overall average quality score $S_{\mathrm{sum}}(\tilde{Y}) = S_{\text{com}}(\tilde{Y}) + S_{\text{con}}(\tilde{Y})$, where $S_{\text{com}}(\tilde{Y})$ and $S_{\text{con}}(\tilde{Y})$ are the completeness and conciseness scores, respectively.
In general, we can define $S_{\mathrm{sum}}(\tilde{Y})$ as the sum or average of a given collection of model-based scores of interest.
For example, we could also have considered a faithfulness score. 
It is worth noting that different quality dimensions often involve trade-offs, {\em i.e.}, improving conciseness may come at the expense of completeness, and {\em vice versa}.
This is justified because these two scores are inherently at odds: {\em striving for completeness often leads to longer summaries with more information, which can reduce conciseness, whereas optimizing for conciseness typically requires discarding certain details, potentially harming completeness}.
Consequently, it is difficult to produce a single summary \( \tilde{Y} \) that maximizes both quality scores simultaneously.
This trade-off arises from the limited word ``budget'' available for a summary, thus compressing content inevitably forces choices about what to include and what to omit, making it challenging to excel on all quality metrics of the summary at once.
Thus, it is desirable to equip \( p_\theta \) with a control mechanism that enables it to generate summaries tailored to different dimensions, {\em e.g.}, completeness or conciseness.
Importantly, \( Z \) in \eqref{eq:autoregressive} now not only guides the summarization process but also controls the specific quality dimension.
Hence, another goal is to guide the model \( p_\theta \) to generate a summary \( \tilde{Y} \) prioritizing a specific criterion $D=\{\text{com},\text{con}\}$, while achieving a high-quality score \( S_{D}(\tilde{Y}) \).

%

\noindent{\bf Optimizing Completeness and Conciseness}\label{sec:oprimize_quality}
Based on the benchmark results using FineSurE reported by \citet{lee2024unisumeval}, existing models exhibit reasonably strong performance in terms of faithfulness but lag behind in completeness and conciseness.
\citet{song2024learning} use Direct Preference Optimization (DPO) \citep{rafailov2024direct} to align large language models with human feedback.
However, DPO is computationally expensive, provided that one needs to compute losses from two models (the policy network and the reference network).
Moreover, \citet{liu2025learning} show that DPO performs poorly on tasks such as ranking, due to its simplistic pairwise ranking logic.
Motivated by these observations, we consider adopting the margin ranking (MR) loss \citep{liu2023learning, chern2023improving, liu2022brio}:
\begin{equation}
\mathcal{L}_{\mathrm{MR}} = \textstyle{\sum}_{k=1}^{K} \sum_{j > k}^{K} \max(0, s_j - s_k + \lambda_{jk})\,,
\label{eq:constra}
\end{equation}
where $\{s_k\}_{k=1}^K$ are the log-likelihood of a collection of $K$ summaries $\{\tilde{Y}_k\}_{k=1}^K$, defined as $s_k = \frac{1}{N} \sum_{i=1}^{N} \log p_\theta(y_i | y_{<i}, X, Z)$, and sorted so that $s_k > s_j$ if $S_{\mathrm{sum}}(\tilde{Y}_k) > S_{\mathrm{sum}}(\tilde{Y}_j)$ for all $k, j \in \{1, \ldots, K\}$.
We set the margin $\lambda_{jk} = \lambda \times (j - k)$ for a hyperparameter $\lambda$, which is selected through cross-validation in our experiments.

Conceptually, the loss in \eqref{eq:constra} encourages the model to produce outputs whose log-likelihood scores are ordered consistently with the model-based quality scores $\{S_{\mathrm{sum}}(\tilde{Y}_k)\}_{k=1}^K$, penalizing pairwise order violations. Moreover, the margin enforces a minimum separation between outputs, improving robustness as supported by the literature on maximum-margin methods~\citep{smola2000advances}.

An unintended consequence of the loss in \eqref{eq:constra} is that it focuses only on preserving the order of $\{S_{\mathrm{sum}}(\tilde{Y}_k)\}_{k=1}^K$, without considering their quality.
To address this and following \citet{liu2025learning}, we introduce an additional objective that directly encourages maximizing the model-based score of the generated summary $\tilde{Y}$ relative to the best reference summary from $\{\tilde{Y}_k\}_{k=1}^K$.
We define
\begin{equation}
\scalebox{0.85}{$
    \mathcal{L}_{\mathrm{MS}} = \max\left(0, \left( S_{\mathrm{sum}}(Y_{\mathrm{ref}}) - S_{\mathrm{sum}}(\tilde{Y}) \right) f(s_1) \right)\,,$}
    \label{eq:bs}
\end{equation}
where $s_1$ is the log-likelihood corresponding to the ML $\tilde{Y}$ under $p_\theta(Y | X, Z)$, $f(\cdot)$ is the exponential function (used to transform log-likelihood to likelihood), and $Y_{\mathrm{ref}}$ is the reference summary taken from the set $\{\tilde{Y}_k\}_{k=1}^K$, which is one with the highest model-based score among $S_{\mathrm{sum}}(\{\tilde{Y}_k\}_{k=1}^K)$.

\begin{figure*}[t]
    \centering
    \includegraphics[width=2\columnwidth]{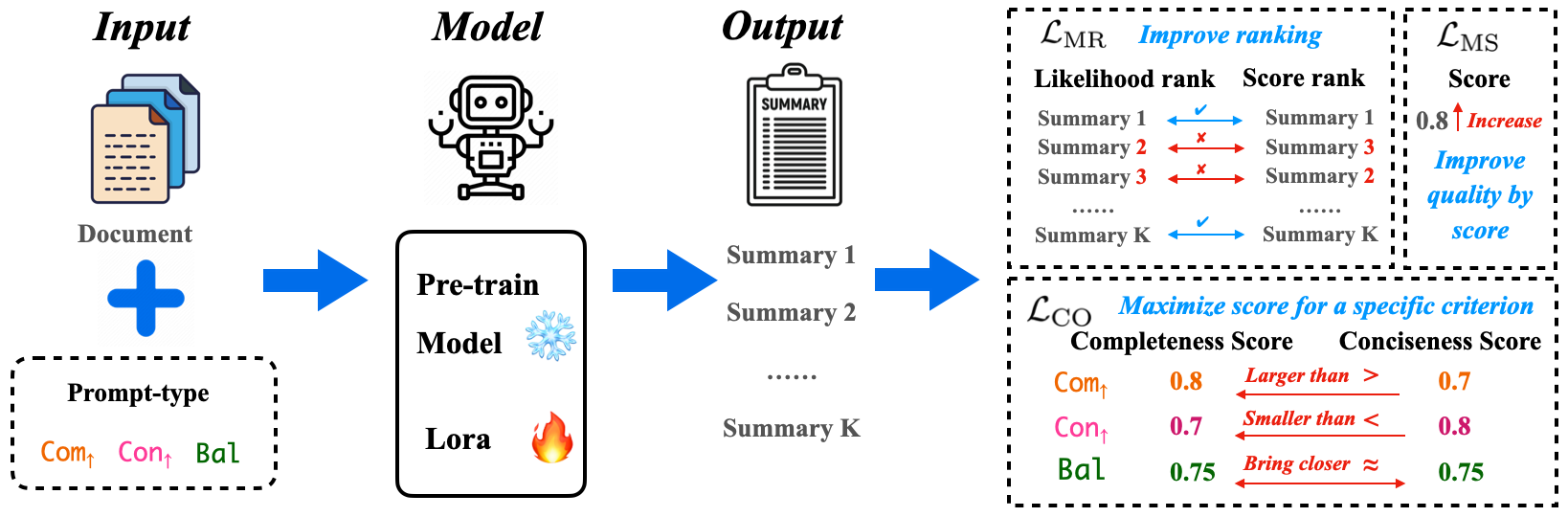}
    \vspace{-3mm}
    \caption{\small
    Model architecture and loss functions.
    The model takes an input document along with a specific prompt to generates $K$ summaries, which are used to compute different loss components.
    The prompts for \textcolor[HTML]{ff7f0e}{Com$_\uparrow$}, \textcolor[HTML]{E22B98}{Con$_\uparrow$} and \textcolor[HTML]{009901}{Bal} are designed to prioritize a more complete, more concise or a balanced summary, respectively.
    }
    \label{fig:architecture}
    \vspace{-3mm}
\end{figure*}

Recall that $\tilde{Y}$ is generated using nucleus sampling.
Thus, $\mathcal{L}_{\mathrm{MS}}$ aims to improve the model-based quality score of the top prediction relative to the best reference.
As shown in Figure~\ref{fig:architecture}, the MR loss $\mathcal{L}_{\mathrm{MR}}$ encourages the alignment of log-likelihoods with model-based scores for a set of $K$ generated summaries, which in the figure is indicated in blue and red for correct and incorrect alignments, respectively.
Moreover, the MS loss $\mathcal{L}_{\mathrm{MS}}$ encourages the model to assign higher scores to the top prediction, thereby improving summary quality overall.

So far, the objective in \eqref{eq:constra} and \eqref{eq:bs} is to produce high-quality summaries.
Next, we propose a learning strategy to train a model capable of generating controllable summaries.

\noindent{\bf Optimizing Controllability}
\citet{song2024learning} examine how focusing on a specific dimension affects summary quality compared to considering all dimensions equally. 
Their experiments shed light on the concept of ``alignment tax'' \citep{noukhovitch2023language, guo2024controllable}, which refers to the trade-off where enhancing alignment with one objective ({\em e.g.}, conciseness) may reduce performance in another ({\em e.g.}, completeness).
However, existing approaches do not effectively control the summary quality along specific dimensions ({\em e.g.}, completeness or conciseness).
To address this limitation, we incorporate the ratio between completeness and conciseness scores as a training signal, guiding the model to shift the summary toward the desired summarization dimensions during training.

For simplicity, we categorize the generation process into three scenarios:
$i$) prioritizing completeness (Com$_\uparrow$),
$ii$) prioritizing conciseness (Con$_\uparrow$), and
$iii$) balancing completeness and conciseness (Bal).
The prompt \( Z \) in \eqref{eq:autoregressive} thus not only guides the summary generation, but also specifies the desired quality trade-off.
Details of the prompts used for each scenario are provided in Appendix~\ref{appendix_control_prompt}. 

To capture these three scenarios, we first define
$S_{\mathrm{ratio}}(\tilde{Y}) = S_{\mathrm{com}}(\tilde{Y})/S_{\mathrm{con}}(\tilde{Y})$.
For scenario $i$), we maximize \( S_{\mathrm{ratio}}(\tilde{Y}) \);
for scenario $ii$), we maximize \( S_{\mathrm{ratio}}(\tilde{Y})^{-1} \); and
for scenario $iii$), we aim for \( S_{\mathrm{ratio}}(\tilde{Y}) \to 1 \). 
We define the control-oriented loss as follows.
%
\vspace{-0.5cm}

\begin{equation}
\hspace{-2mm}
\scalebox{0.64}{$
\mathcal{L}_{\mathrm{CO}} =
\begin{cases}
\max\!\bigl(0,\,[ S_{\mathrm{ratio}}(Y_{\mathrm{ref}})-S_{\mathrm{ratio}}(\tilde{Y}) ] f(s_1) \bigr), 
 & \text{Com$_\uparrow$}, \\[0.6ex]
\max\!\bigl(0,\,[ S_{\mathrm{ratio}}(\tilde{Y})-S_{\mathrm{ratio}}(Y_{\mathrm{ref}}) ] f(s_1) \bigr), 
 & \text{Con$_\uparrow$}, \\[0.6ex]
\max\!\bigl(0,\,[ |\log S_{\mathrm{ratio}}(\tilde{Y})|-|\log S_{\mathrm{ratio}}(Y_{\mathrm{ref}})| ] f(s_1) \bigr), 
 & \text{Bal} ,
\end{cases}
$}
\label{eq:co}
\end{equation}
where \( s_1 \) is the log-likelihood of the ML generated summary \( \tilde{Y} \), $f(\cdot)$ is the exponential function (used to transform log-likelihood to likelihood), and $Y_{\mathrm{ref}}$ is the reference summary selected from the set $\{\tilde{Y}_k\}_{k=1}^K$, which is the one with the highest model-based score among $S_{\mathrm{ratio}}(\{\tilde{Y}_k\}_{k=1}^K)$.
This loss enforces the desired adjustment in the ratio $ S_{\mathrm{ratio}}(\tilde{Y})$, pushing it higher when completeness is prioritized, lower when conciseness is prioritized, and closer to 1 in the balanced case.

As shown in Figure~\ref{fig:architecture}, the control loss $\mathcal{L}_{\rm CO}$ introduces prompt-specific control signals.
Input prompts are categorized into one of three types: Com$_{\uparrow}$, Con$_{\uparrow}$  or Bal.
The model learns to condition its output on these control types, favoring completeness over conciseness for the first, conciseness over completeness for the second, and balancing both for the third.
This enables fine-grained control over summary characteristics, improving flexibility and alignment with user intent.

Although $\mathcal{L}_{\rm CO}$ enables the model to generate summaries aligned with prompt-specific preferences, the scoring mechanism in \eqref{eq:constra} applies a uniform aggregation of completeness and conciseness scores, which lacks the granularity needed to reflect these nuanced control signals.
In \eqref{eq:constra}, the summaries are ranked according to their overall score \(S_{\mathrm{sum}}(\tilde{Y}) = S_{\mathrm{com}}(\tilde{Y}) + S_{\mathrm{con}}(\tilde{Y})\).
However, this simple aggregation fails to reflect fine-grained preferences.
For example, summaries with \((S_{\mathrm{com}}=0.9, S_{\mathrm{con}}=0.1)\), \((S_{\mathrm{com}}=0.1, S_{\mathrm{con}}=0.9)\), and \((S_{\mathrm{com}}=S_{\mathrm{con}}=0.5)\) produce the same \(S_{\mathrm{sum}}\), but exhibit very different trade-offs.
So motivated, we update \(S_{\mathrm{sum}}\to S_{\mathrm{sum*}}\) in \eqref{eq:constra} to incorporate scenario-specific penalties.
Specifically,
\begin{equation}
S_{\mathrm{sum*}} = S_{\mathrm{sum}} - \delta \cdot \phi(S_{\mathrm{com}}, S_{\mathrm{con}}),
\label{eq:MR_term2}
\end{equation}
with the penalty term \(\phi(\cdot,\cdot)\) defined as:
%
\[
\scalebox{0.75}{$
\phi(S_{\mathrm{com}}, S_{\mathrm{con}}) =
\begin{cases}
S_{\mathrm{con}} - S_{\mathrm{com}}, & \text{Com$_\uparrow$}, \\
S_{\mathrm{com}} - S_{\mathrm{con}}, & \text{Con$_\uparrow$}, \\
|S_{\mathrm{com}} - S_{\mathrm{con}}|, & \text{Bal},
\end{cases}
$}
\]
where \(\delta > 0\) is a hyperparameter (set using cross-validation) that determines the strength of the penalty for deviation from the desired trade-off.
This loss encourages fine-grained priorities to better align with the specified criteria when the average scores of completeness and conciseness are identical across options.
By subtracting $\delta\cdot\phi$ from $S_{\mathrm{sum}}$, $S_{\mathrm{sum*}}$ becomes a more discriminative scoring function that aligns better with task-specific preferences, even when raw scores $S_{\mathrm{sum}}$ are tied.

Finally, we combine the margin ranking loss in \eqref{eq:constra} modified using \eqref{eq:MR_term2} with the score-improving loss in \eqref{eq:bs} and control-oriented loss \eqref{eq:co} as
\begin{equation}
\mathcal{L}_{\rm Total} =\mathcal{L}_{\rm MR} + \gamma\mathcal{L}_{\rm MS}+\beta \mathcal{L}_{\rm CO} ,
\end{equation}
where $\gamma$ is a hyperparameter that balances the improvement of model-based scores and $\beta$ balances controllability.
These $\{\gamma,\beta\}$ are tuned using cross-validation in our experiments.
{Details of the FineSurE calculation and the corresponding prompt designs can be found in Appendices~\ref{app:finesure} and~\ref{appendix:prompt_finesure}.}

\section{Experiments}\label{section:experiment}
{\bf Datasets}
We consider three datasets from five different domains.
The FeedSum dataset \citep{song2024learning} includes separate training and test sets drawn from different domains of documents.
The training set contains a diverse collection of model-generated summaries, each annotated with fine-grained FineSurE scores.
Due to computational constraints, we focus on short document types in this study.
Specifically, we use three in-domain datasets for training and evaluation, namely Wikihow (lifestyle)~\citep{koupaee2018wikihow}, CNN/DM (news)~\citep{nallapati2016abstractive}, and DialogSum (daily life conversations)~\citep{chen2021dialogsum}.
We also evaluate generalization to two out-of-domain summarization datasets: OpoSum (product reviews)~\citep{angelidis2018summarizing} and MeQSum (medical)~\citep{abacha2019summarization}.
In the Appendix, we show Table~\ref{tab:dataset_summary}, which summarizes these three datasets.

\noindent{\bf Baselines}
We compare our method with five popular prompt-based LLMs: LLaMA~\citep{touvron2023llama, touvron2023llama2, dubey2024llama}, Qwen~\citep{yang2024qwen2}, Mistral~\citep{jiang2023mistral7b}, GPT~\citep{ouyang2022training} and Gemini~\citep{team2023gemini}.
For LLaMA, we consider two variants: \texttt{LLaMA-3.1-8B-Instruct} and \texttt{SummLLaMA3-8B}, the latter being a fine-tuned version trained on the FeedSum dataset~\citep{song2024learning}.
For Qwen, we use the \texttt{Qwen2.5-7B-Instruct} model, and for Mistral, we evaluate \texttt{Mistral-7B-Instruct}.
For GPT, we consider \texttt{GPT-4o}, \texttt{GPT-4o-mini}, and \texttt{GPT-4-turbo}.
For Gemini, we only use \texttt{gemini-2.0-flash}.
The prompts for controlled summary generation can be found in Appendix~\ref{appendix_control_prompt}.

\begin{figure*}[t]
    \centering
    \includegraphics[width=450pt]{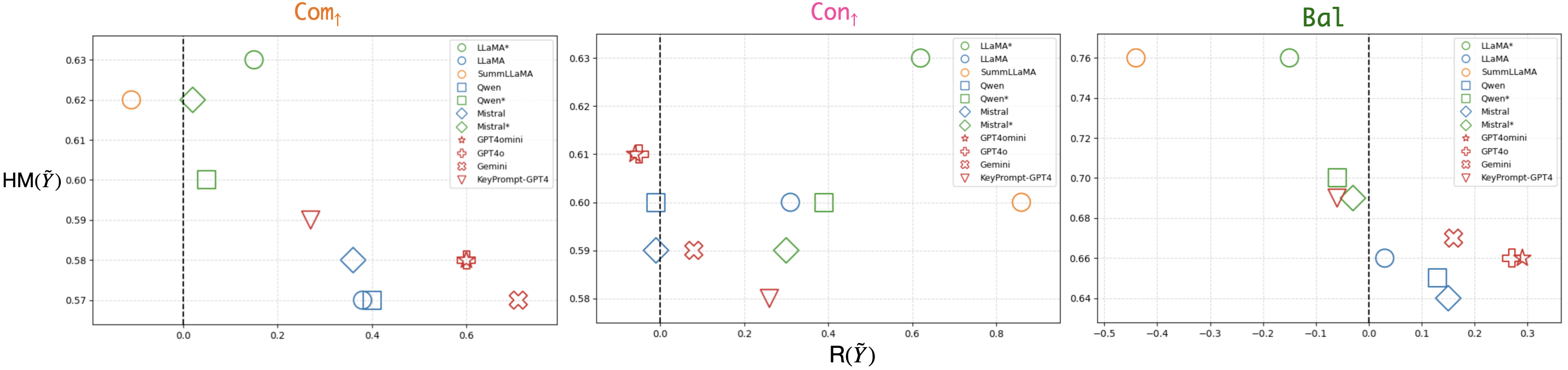}
    \vspace{-5pt}
    \caption{\small Model performance across different control settings. Each point represents the mean of $\text{HM}(\tilde{Y})$ (y-axis) and  $\text{R}(\tilde{Y})$ (x-axis) across all test cases in FeedSum test set. Models are grouped by color into four categories: \textcolor[HTML]{1f77b4}{Baseline models (blue)}, \textcolor[HTML]{2ca02c}{Our methods (green)}, \textcolor[HTML]{FFA500}{SummLLaMA (orange)}, and \textcolor[HTML]{d62728}{Commercial models (red)}. The three panels show performance under different control priorities: \textcolor[HTML]{ff7f0e}{Com$_\uparrow$} prioritizes completeness, \textcolor[HTML]{E22B98}{Con$_\uparrow$}  prioritizes conciseness, and \textcolor[HTML]{009901}{Bal} aims to balance both. The vertical dashed line at  $\text{R}(\tilde{Y})$ = 0 represents the controllability target (reference) between completeness and conciseness.}
    \label{fig:main_comparison}
    \vspace{-5mm}
\end{figure*}

\noindent{\bf Evaluation Metrics}
We use Spearman's rank correlation coefficient to measure the alignment between model predictions $\log p_{\theta}(\{\tilde{Y}_k\}_{k=1}^K$ and model-based scores $S_{\mathrm{sum}}(\{\tilde{Y}_k\}_{k=1}^K)$.
A higher Spearman correlation indicates stronger alignment between the model's likelihood estimates and the model-based scores, suggesting that the model assigns higher probabilities to summaries that are rated more favorably.
We also consider the harmonic mean of completeness and conciseness scores to evaluate the overall quality of a summary: $\text{HM}(\tilde{Y}) = 2/(S_{\mathrm{com}}(\tilde{Y})^{-1} + S_{\mathrm{con}}(\tilde{Y})^{-1})$.
A higher harmonic mean $\text{HM}(\tilde{Y})$ is associated with better overall summary quality.
Lastly, we obtain the control ratio to assess the ability of the model to guide summaries toward the desired dimension: $\text{R}(\tilde{Y}) = ( \log( S_{\mathrm{com}}(\tilde{Y})/S_{\mathrm{con}}(\tilde{Y})) )^{\alpha}$, where $\alpha = 1$ when prioritizing completeness or balance; and $\alpha = -1$ when prioritizing conciseness.
We expect $\text{R}(\tilde{Y})$ to be high when completeness or conciseness is explicitly prioritized, depending on the value of $\alpha$, and to be close to zero when balance is desired.

\noindent{\bf Implementation}
Our methodology considers \texttt{Qwen2.5-7B-Instruct}, \texttt{Mistral-7B-Instruct}, and \texttt{LLaMA-3.1-8B-Instruct} using a single NVIDIA H100 GPU.
To increase the diversity of the candidate pool, we obtain $K$-candidate summaries by combining $K-1$ reference summaries randomly sampled from the FeedSum dataset \citep{song2024learning}, which contains summaries generated by various models, with the top summary prediction produced by our model via nucleus sampling.
These summaries constitute the set of candidates used to compute the loss.
The model is fine-tuned for three epochs, and during each iteration, the newly generated summary is added to the reference summary pool.
All models were initialized from pre-trained backbones and fine-tuned using LoRA~\citep{hu2022lora}. 
The hyperparameter selection and their settings are in Appendix~\ref{appendix:hyperparams}.

For ranking-based evaluation, we set our own test split from the FeedSum training data since the original FeedSum~\citep{song2024learning} test set does not contain diverse model-generated summaries with corresponding FineSurE scores.
Specifically, we sampled 100 summaries from each domain to form a test set for ranking-based evaluation, and used the remaining 12,000 examples as the training set.
The original FeedSum test set was retrained for summary quality evaluation using $\text{HM}(\tilde{Y})$, and control ability using $\text{R}(\tilde{Y})$ and FineSurE scores obtained through GPT-4o, which was selected to align with SummLLaMA \citep{song2024learning}, and thus ensuring a consistent comparison.


\noindent{\bf Ranking performance}
Section~\ref{sec:oprimize_quality} introduced our approach to aligning model predictions with model-based scores, specifically using the margin-ranking objective (${\cal L}_{\rm MR}$) in \eqref{eq:constra}. Table~\ref{tab:full_model_spea} shows the results using (median) Spearman correlations to compare alignment quality.
These are reported for all cases in the FeedSum test dataset.
For completeness, we also report the Pearson and Kendall’s tau correlation metrics in the Appendix Table~\ref{tab:pearson_kendall}.

\begin{table}[t]
\centering
\small{
\begin{tabular}{l r c}
\textbf{Model} & \textbf{Median} & \textbf{IQR}\\
\hline
LLaMA        & 0& (-0.3, 0.22)\\
SummLLaMA    & 0.04  &(-0.26, 0.38)\\
LLaMA*     & 0.15& (-0.19, 0.37)\\
Qwen        & -0.02 &(-0.34, 0.29)\\
Qwen*        & \textbf{0.22} & (-0.02, 0.46)\\
Mistral    & -0.03 &(-0.23, 0.19)\\
Mistral*   & 0.09 & (-0.19, 0.38)\\
\end{tabular}}
\vspace{-1mm}
\caption{\small Spearman correlation between model likelihood and model-based scores for $K$ summaries and all cases in the test dataset.
We report the median and interquartile range (IQR).
The best result is highlighted in \textbf{bold}.
The asterisk (*) indicates the base foundation model fine-tuned using the proposed method.}
\label{tab:full_model_spea}
\end{table}

Among all models evaluated, Qwen* achieves the highest median Spearman correlation (0.22), with an interquartile range of (-0.02, 0.46).
Qwen* refers to the Qwen base model fine-tuned using the proposed method. 
This result suggests that our alignment approach is effective when applied to the Qwen architecture.
LLaMA*, ranks second with a median score of 0.15 and a relatively tight IQR of (-0.19, 0.37), which outperforms its base version (LLaMA, median 0.00) and the variant SummLLaMA (median 0.04), further demonstrating the general effectiveness of our training strategy in enhancing alignment with quality scores.
The consistently higher Spearman correlations for all our fine-tuned models indicate that our method may also be useful for summary quality ranking via model likelihoods.
The Spearman scores distributions for all models are shown in Appendix Figure~\ref{fig:ranking_distribution}.

\begin{figure*}[t]
    \includegraphics[width=0.8\textwidth]{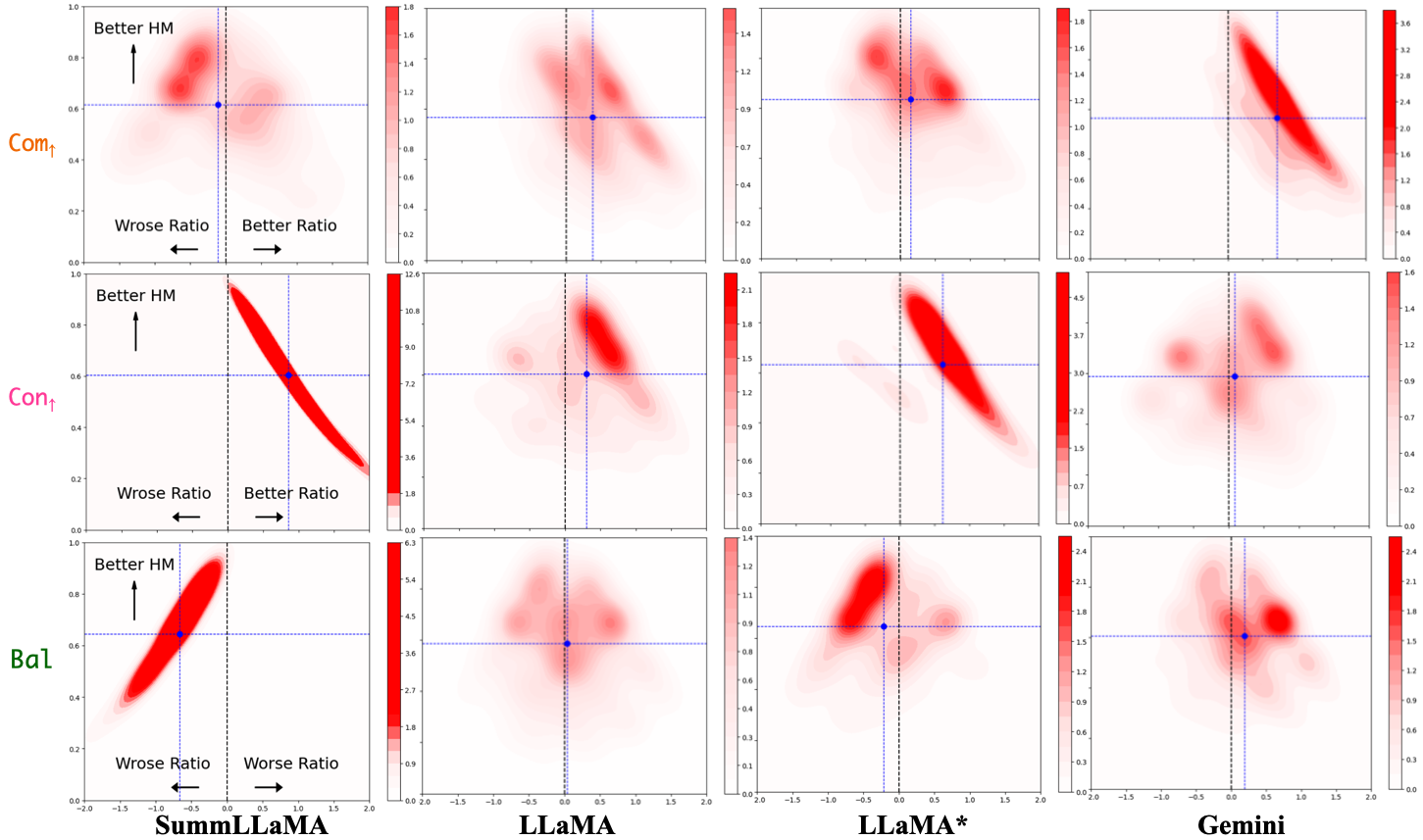}
    \vspace{-2mm}
    \centering
    \caption{\small Distributions of  $\text{R}(\tilde{Y})$ (x) and $\text{HM}(\tilde{Y})$ (y) metrics.
    \textcolor[HTML]{ff7f0e}{Com$_\uparrow$} prioritizes completeness, \textcolor[HTML]{E22B98}{Con$_\uparrow$}  prioritizes conciseness, and \textcolor[HTML]{009901}{Bal} to balance them.
    \textcolor[HTML]{0000FF}{Blue dashed lines}
     mark the mean of the metrics, and the arrows point in the direction in which metrics are better or worse.}
    \label{fig:contour_main}
    \vspace{-6mm}
\end{figure*}

\noindent{\bf Quality and Controllability Performance}
In \eqref{eq:bs} (${\cal L}_{\rm MS}$), we introduced a loss to improve the quality of model-generated summaries.
We now evaluate the quality of these summaries using FineSurE.
Specifically, we seek high completeness {\em and} conciseness scores for generated summaries, which we evaluate using the $\text{HM}(\tilde{Y})$ score introduced above.
%
Complementary, for controllability performance we seek high $\text{R}(\tilde{Y})$ values for Com$_\uparrow$ and Con$_\uparrow$, which imply that the model assigns greater importance to the higher-priority attribute, completeness or conciseness, respectively; or $\text{R}(\tilde{Y})$ $\to 0$ when a balanced (Bal) summary is of interest.
In general, the objective is to maintain high quality generation while ensuring effective control.
Therefore, a higher $\text{HM}(\tilde{Y})$ value is preferred, along with preferably a higher $\text{R}(\tilde{Y})$. 

Figure~\ref{fig:main_comparison} separately shows the mean $\text{HM}(\tilde{Y})$ and $\text{R}(\tilde{Y})$ metrics for the three control scenarios (Com$_\uparrow$, Con$_\uparrow$ and Bal), 10 different methods (3 baseline open source models, 3 corresponding fine-tuned models, SummLLaMa and three commercial models), for all the test cases in the FeedSum test set.
%
We observe the following:
$i$) In the (Com$_{\uparrow}$) and (Con$_{\uparrow}$) scenarios, we seek models on the right-hand side of the $\text{R}(\tilde{Y})$ scale ({\em i.e.}, with  $\text{R}(\tilde{Y})>0$) while also producing high $\text{HM}(\tilde{Y})$ values.
However, SummLLaMA (orange) achieves reasonable $\text{HM}(\tilde{Y})$ scores but consistently produces negative $\text{R}(\tilde{Y})$ values in the (Com$_{\uparrow}$) setting, favoring conciseness regardless of the intended control.
This indicates limited controllability, which is likely due to its optimization for general summarization quality rather than controllable generation.
$ii$) Our models (green), especially LLaMA*, achieve high $\text{HM}(\tilde{Y})$ scores and positive $\text{R}(\tilde{Y})$ values, demonstrating effective controllability and strong summary quality. 
$iii$) In the (Bal) setting, we aim for models near the dashed line ($\text{R}(\tilde{Y})$$\to 0$), indicating balanced attention to both completeness and conciseness, together with high $\text{HM}(\tilde{Y})$.
Here, LLaMA* matches SummLLaMA in $\text{HM}(\tilde{Y})$ but is much closer to the target $\text{R}(\tilde{Y})$, showing that our method achieves better controllability without sacrificing quality.
$iv$) Baseline models (blue) generally exhibit higher controllability, but underperform in terms of $\text{HM}(\tilde{Y})$ compared to our methods.
$v$) Commercial models (red) exhibit good controllability ( $\text{R}(\tilde{Y})$ aligned with the desired direction), but their overall $\text{HM}(\tilde{Y})$ scores are lower. $vi$) Keyprompt is a prompt-oriented baseline in which we prompt the model to first generate key facts and then produce a summary conditioned on these key facts.
Although the prompt-based method offers flexibility and achieves reasonable controllability ({\em i.e.}, $\text{R}(\tilde{Y})$), it comes at a significant cost in terms of overall summary quality, with its $\text{HM}(\tilde{Y})$ being considerably lower than our fine-tuned LLaMA* across all control settings.

These results demonstrate that our models consistently achieve strong performance and robust control, outperforming both baselines and commercial systems in balancing quality with controllability.
Detailed values for all baseline model comparisons are provided separately in Appendix Table~\ref{tab:model_full_comparison}.
To better illustrate the ability of each model to generate high-quality summaries (by $\text{HM}(\tilde{Y})$) and controllability (by $\text{R}(\tilde{Y})$), we show their distributions (as contours) in Figure~\ref{fig:contour_main}, where the x and y axes represent $\text{R}(\tilde{Y})$ and $\text{HM}(\tilde{Y})$, respectively, and the color indicates density.
The blue marker is the mean $\text{HM}(\tilde{Y})$ and $\text{R}(\tilde{Y})$ from Table~\ref{tab:model_full_comparison}.
%
We expect distributions to be concentrated in the upper-middle region, reflecting both high summary quality and effective control.
For the scenarios Com$_{\uparrow}$ and Con$_{\uparrow}$, we hope to see the distribution tail extend to the right, which indicates a consistent control bias aligned with the prompt while maintaining high quality.
Under the Bal setting, we hope for a clustered distribution in the upper-middle region, thus summaries that are both complete and concise.

As shown in Figure~\ref{fig:contour_main}, among all models:
$i$) LLaMA* (our method) consistently produces the most favorable distribution, concentrated in the desired region for all three settings, and generally shifted upward (indicating high $\text{HM}(\tilde{Y})$), suggesting both strong control and high-quality output.
$ii$) In contrast, across all settings, SummLLaMA consistently produces overly concise summaries (evidenced by a leftward shift in the Com$_{\uparrow}$ and Bal settings), which indicates poor controllability.
$iii$) LLaMA demonstrates better control alignment ({\em i.e.}, correct shift in $\text{R}(\tilde{Y})$), but its lower $\text{HM}(\tilde{Y})$ values suggest limited summary quality.
Results for all models are shown in Appendix Figures~\ref{fig:appendix_1} and~\ref{fig:appendix_contour_openai}.

\begin{table}[th]
\centering
\scalebox{0.8}{
\begin{tabular}{lcc}
\textbf{Model} & \textbf{Median} & \textbf{IQR} \\
\hline
LLaMA* (MR)         & 0.07 & ($-0.27$, $0.37$) \\
LLaMA* (MS)         & 0.04 & ($-0.26$, $0.38$) \\
LLaMA* (CO)         & 0.05 & ($-0.26$, $0.38$) \\
LLaMA* (MR+MS)      & 0.14 & ($-0.16$, $0.38$) \\
LLaMA* (MR+CO)      & 0.12 & ($-0.16$, $0.36$) \\
LLaMA* (MS+CO)      & 0.04 & ($-0.26$, $0.38$) \\
LLaMA* (MR+MS+CO)   & \textbf{0.15} & ($-0.19$, $0.37$) \\
\end{tabular}}
\caption{Spearman correlation analysis for ablation on LLaMA loss components. The best results are highlighted in bold, and the asterisk ($*$) denotes the base LLaMA model fine-tuned using the proposed objective.}
\vspace{-2mm}
\label{tab:ablation_rank}
\end{table}

\noindent{\bf Ablation study}  
We conducted an ablation study to assess the individual contributions of each component in our loss function: MR (ranking), MS (generation quality), and CO (control), all under the same training setup.

In Table~\ref{tab:ablation_rank}, we compute the Spearman correlation between model likelihoods and model-based evaluation scores over $K$ sampled summaries for all test cases, under different LLaMA fine-tuning settings with ablated loss components. This evaluates how well the model aligns with summary quality. We report the median and interquartile range (IQR) for each configuration.  Using MR alone yields only limited improvements in ranking performance.
However, when MR is combined with MS or CO, the ranking results improve noticeably.
In fact, the full combination of MR, MS, and CO achieves the highest median correlation, indicating a strong synergy among the losses.
For control quality, in Appendix Table~\ref{tab:ablation_finesure}, we show that incorporating MS leads to better generation quality, reflected in higher $\text{HM}(\tilde{Y})$ scores, while adding CO enhances controllability, as shown by improvements in $\text{R}(\tilde{Y})$.
The combination of the three losses results in a well-balanced trade-off between summary quality and controllability.
Thus, these findings highlight the complementary roles of the loss components and provide empirical support for the effectiveness of our composite training objective.

\noindent{\bf Human Study}
To assess whether our model effectively learns to generate summaries with controllable attributes, we conducted a human evaluation based on pairwise comparisons.

We use our fine-tuned model (LLaMA*) to generate summaries under different control prompts (Com$_{\uparrow}$, Bal, Con$_{\uparrow}$) and construct pairwise comparisons between them.
To reduce potential biases related to length, we filtered for summary pairs in which both outputs contained the same number of sentences.
From this filtered set, we sampled 60 summary pairs for human annotation.
These were divided into five sets of annotations, each containing 20 pairs.
The first 10 pairs were shared among all annotators to measure inter-annotator agreement (IAA), while the remaining 10 were unique to each annotator.
The evaluation of the human-model alignment used all pairs.
For this task, annotators were asked to read two summaries generated from the same document and select the one they considered more complete.
No additional instructions on conciseness or fluency were provided, as the focus was solely on completeness.
Then, their preferences were compared to the control signal given to the model.
We assumed an ordinal relationship among the control settings (Com$_{\uparrow}$ > Bal > Con${_\uparrow}$), and based on this, assigned a pseudo-label to each pair, {\em e.g.}, if summary A was generated under a higher completeness setting than summary B, it was labeled 1, indicating that A should be more complete; otherwise, labeled 0.

\begin{table}[t]
\centering
\small
\scalebox{1}{
\begin{tabular}{cccccc}
 & \bf{A\_1} & \bf{A\_2} & \bf{A\_3} & \bf{A\_4} & \bf{A\_5} \\
\hline
A\_1 & 1.00 & 0.60 & 0.80 & 0.60 & 1.00 \\
A\_2 & -   & 1.00 & 0.40 & 0.60 & 0.60 \\
A\_3 & -   & -   & 1.00 & 0.40 & 0.80 \\
A\_4 & -   & -   & -   & 1.00 & 0.60 \\
A\_5 & -   & -   & -   & -   & 1.00 \\[0.2ex]
\hdashline
Average $\pm$ Std & \multicolumn{5}{c}{0.64 $\pm$ 0.18}
\end{tabular}}
\vspace{-2mm}
\caption{Inter-annotator agreement matrix for Cohen’s kappa ($\kappa$) with average and standard deviation.}
\label{app:human_study_iaa}
\vspace{-1mm}
\end{table}

\begin{table}[t]
\centering
\small
\scalebox{1}{
\begin{tabular}{lcc}
\bf{Annotator} & \bf{Accuracy} & \bf{Spearman ($\rho$)} \\
\hline
A\_1 & 0.95 & 0.903 \\
A\_2 & 0.80 & 0.612 \\
A\_3 & 0.65 & 0.385 \\
A\_4 & 0.95 & 0.903 \\
A\_5 & 0.90 & 0.816 \\[0.2ex]
\hdashline
Average $\pm$ Std & \textbf{0.85 $\pm$ 0.18} & \textbf{0.72 $\pm$ 0.22} \\
\end{tabular}}
\vspace{-2mm}
\caption{Annotator performance in terms of Accuracy and Spearman correlation ($\rho$).}
\label{app:human_study_accc}
\vspace{-5mm}
\end{table}

Five annotators participated in the study.
IAA was assessed using Cohen’s kappa ($\kappa$) on the 10 shared examples.
Table~\ref{app:human_study_iaa} shows an average $\kappa=0.64\pm0.18$, indicating moderate to substantial agreement and consistent annotation behavior among annotators.
To evaluate model-human alignment, we computed the accuracy and Spearman's rank correlation ($\rho$) between the model's pseudo-labels and human preferences, using all 20 examples per annotator.
Table~\ref{app:human_study_accc} reports an average accuracy of $0.85\pm0.18$ and an average Spearman's of $\rho=0.724\pm0.22$.
These results demonstrate a strong alignment between the model's control-guided outputs and human judgments, supporting the effectiveness of our approach for controllable summary generation.

\noindent{\bf Domain-Specific Results}  
To assess domain adaptation capabilities, we analyze the performance on WikiHow (lifestyle), CNN/DM (news), and DialogSum (dialog), as shown in Appendix Figure~\ref{fig:in_domain}.
We observe the following:
$i$) Our model achieves strong $\text{HM}(\tilde{Y})$ and $\text{R}(\tilde{Y})$ scores in all domains.
However, in the (Com${\uparrow}$) setting on CNN/DM, models such as Qwen* and Mistral* produce negative $\text{R}(\tilde{Y})$, indicating the difficulty in generating complete summaries.
$ii$) In contrast, commercial models consistently achieve positive $\text{R}(\tilde{Y})$ under the (Com${\uparrow}$) setting across all domains, demonstrating their effectiveness in prioritizing completeness.
$iii$) SummLLaMA consistently exhibits negative $\text{R}(\tilde{Y})$ in both (Com$_{\uparrow}$) and (Bal) settings, despite achieving high $\text{HM}(\tilde{Y})$ scores, highlighting a lack of controllability across domains.
To evaluate out-of-domain robustness, we assess performance in OpoSum (product reviews) and MeQSum (medical summaries), as shown in Appendix Figure~\ref{fig:oo_domain}.
Similar trends are observed:
$i$) Commercial models lead in both $\text{HM}(\tilde{Y})$ and $\text{R}(\tilde{Y})$, indicating that open-source models still have room for improvement.  
$ii$) our method outperforms the baselines in both $\text{HM}(\tilde{Y})$ and controllability, and SummLLaMA remains overly concise.
The complete results are provided in Appendix Tables~\ref{tb:in_domain} and~\ref{tb:ood}.

\noindent{\bf Extended Quality Evaluation}
To prevent {\em reward hacking}~\citep{skalse2022defining}, we evaluate models using G-Eval+~\citep{lee2024unisumeval}, as shown in Appendix Figure~\ref{fig:g-eval}.
We observe the following:
$i$) Nearly all models exhibit negative $\text{R}(\tilde{Y})$ values under the Com$_{\uparrow}$ and Bal settings, indicating limited controllability. Our method mitigates this issue under the LLaMA and Qwen frameworks, shifting the  $\text{R}(\tilde{Y})$ toward the positive direction while also improving $\text{HM}(\tilde{Y})$.
$ii$) LLaMA* consistently outperforms both LLaMA and SummLLaMA in terms of overall quality ($\text{HM}(\tilde{Y})$) and controllability ($\text{R}(\tilde{Y})$), demonstrating a better balance between the two.
$iii$) Commercial models achieve strong performance, with consistently high scores in both $\text{HM}(\tilde{Y})$ and $\text{R}(\tilde{Y})$, reflecting superior summary quality and controllability.
The complete results are in Appendix Table~\ref{tb:G-Eval}.
Moreover, we considered additional dimensions (consistency, coherence, fluency and relevance) using G-Eval~\citep{liu2023g}, and ROUGE~\citep{lin2004rouge} and BERTScore~\citep{zhang2019bertscore} compared against human reference summaries.
The results in Appendix Table~\ref{tab:g_eval} show that our model (LLaMA*) consistently outperforms the SummLLaMA baseline in most G-Eval metrics and shows clear improvements over untrained LLaMA, demonstrating the effectiveness of our control-aware training.
Improvements in ROUGE and BERTScore are comparatively smaller, likely due to a broad overlap with the reference summaries.
We also report the faithfulness score in Table~\ref{tb:faithfulness} and Table~\ref{tb:G-Eval}, showing that our method maintains high faithfulness.

\noindent{\bf Dimension Extension}
To verify the effectiveness of our method, we also conduct experiments along different dimensions. In this setting, we focus on completeness and faithfulness.
As shown in Appendix Figures~\ref{fig:com_fai_comparison} and~\ref{fig:contour_com_fai}, both the baseline model and SummLLaMA struggle to achieve positive performance in the Com$_\uparrow$ setting ($\text{R}(\tilde{Y})>0$).
In contrast, after training with our approach, the model can be effectively guided along the desired dimension.
We report the detailed results in Table~\ref{tab:model_com_fai}.

\vspace{-2mm}
\section{Discussion}\label{sec:discussion}
\vspace{-2mm}
%
Our proposed method is related to Reinforcement Learning from AI Feedback (RLAIF)  in that it learns from AI feedback. However, the key distinction lies in the training objective and computational requirements.
Standard RL approaches, such as PPO~\citep{schulman2017proximal} or GRPO~\citep{shao2024deepseekmath}, require keeping a reference model to compute the policy loss and to constrain policy updates. This design increases both memory consumption and training cost; as it involves additional forward passes and parameter storage.
In contrast, our method adopts a ranking-based objective rather than a reinforcement learning objective. It does not require hosting with a reference model. As a result, the training process is simpler and computationally more efficient.
The current work focuses on two important summarization dimensions, while the proposed framework is readily compatible with other perspectives, such as stylistic attributes, which can be incorporated in future work through the same scoring and ranking formulation.
%

\vspace{-1mm}
\section*{Limitations}
\vspace{-1mm}
The method is heavily dependent on the quality of the external scoring function FineSurE. Although it correlates with human judgments, it remains an imperfect proxy, and thus it can introduce biases or systematic errors. If the scorer fails to capture certain aspects of quality ({\em e.g.}, fluency or discourse coherence), the model may overfit to these incomplete signals, leading to reward misalignment or reward hacking.
The approach also relies on sampling multiple candidate summaries $K$ to compute ranking losses, which introduces additional computational overhead compared to standard maximum likelihood training. Moreover, the training signal is sensitive to the quality and diversity of the sampled candidates.
Finally, the framework is validated only on two quality dimensions ({\em e.g.}, completeness and conciseness). Although the formulation is general, its effectiveness in scenarios involving three or more dimensions remains unexplored.





\vspace{-4mm}
\bibliography{custom}

\clearpage
\appendix


\section{Related Work}\label{app:related}
{\bf Summary Evaluation}
Traditional summarization metrics ({\em e.g.}, ROUGE) correlate poorly with human judgments of semantic fidelity and factuality.
Recent LLM-based evaluators instead assess content at a finer granularity~\citep{laban2023summedits,lu2023toward}.
To be specific, FineSurE~\citep{song2024finesure} extracts atomic \emph{keyfacts} from the source and summary, computing completeness and conciseness by aligning these keyfacts with summary sentences, delivering interpretable and dimension‐specific scores. UniSumEval~\citep{lee2024unisumeval} provides a unified and robust resource for benchmarking summarization systems, facilitating a more accurate and comprehensive performance evaluation.
Collectively, these works mark a shift from token-level to content-level evaluation, paving the way for more reliable and actionable supervision. Our work moves a step further by using these scores as a reward signal for model refinement.

\noindent{\bf Summary Optimization}
Training summarization models directly with non-differentiable metrics has been attempted via reinforcement learning (RL) and preference learning.
Early RL methods treat evaluation scores as rewards~\citep{kaelbling1996reinforcement}, {\em e.g.}, Proximal Policy Optimization (PPO)~\citep{schulman2017proximal} or Direct Preference Optimization (DPO)~\citep{rafailov2024direct}.
However, they come at the cost of either high variance and instability or substantial computation cost.

Ranking-based objectives offer a more scalable alternative.
BRIO~\citep{liu2022brio} casts summarization as a contrastive ranking task, encouraging higher-quality candidates to outrank inferior ones. 
Subsequent work introduces margin ranking losses to enforce explicit quality separations~\citep{liu2023learning, chern2023improving}. 
Contrastive sequence learning approaches like GEC-Sum~\citep{xie2024gecsum} further integrate automated evaluation scores into end-to-end fine-tuning.
More recently, \citet{song2024learning} showed that LLM-generated critiques can serve as effective supervision signals, closing the gap between human feedback and model training. \citet{yun2025refeed} propose an inference-time refinement method that performs multiple iterations for a single input. Given a document, the model first generates an initial summary and then computes its FineSurE score. This score is fed back to the model as a feedback signal to refine the summary, and the process is repeated until the summary reaches a sufficiently high score. 

Despite these advances, most methods optimize a single aggregated score and lack mechanisms to steer generation along distinct quality dimensions.  Importantly, controlling or steering model generation has been increasingly recognized as essential~\citep{wang2025think}. This is further supported by work on the ``alignment tax’’~\citep{noukhovitch2023language, guo2024controllable}, which reveals that improving one objective ({\em e.g.}, conciseness) often degrades another ({\em e.g.}, completeness). 
In contrast, our approach combines fine-grained LLM-based scoring, margin ranking, and a control-oriented loss to enable high-quality summaries that can be flexibly tuned toward completeness, conciseness, or a balanced trade-off.

\section{FineSurE}\label{app:finesure}
The Finegrained Summarization Evaluation (FineSurE) model is a recently proposed model-based score that uses large language models (LLMs) to evaluate the summarization quality of a generator at a fine-grained level through summary sentences they call {\em keyfacts} \citep{song2024finesure}.
A keyfact is defined as a short sentence conveying a single key information element, consisting of at most 2 or 3 entities, where an entity refers to a salient concept, object, or named item ({\em e.g.}, people, organizations, locations, or events).
These keyfacts are also known as {\em semantic content units} \citep{bhandari2020re}.

To measure the completeness and conciseness of a generated summary, we rely on a process called {\em keyfact alignment}, which assesses how well the content of the summary corresponds to the key information elements in the source document.
The alignment of keyfacts serves as the foundation for computing both completeness and conciseness scores.
It involves determining which keyfacts extracted from the source document are present in the summary and then identifying the summary sentences that express them.
Although humans are generally best at generating keyfacts, particularly in specialized domains such as medicine or sales, where content prioritization requires domain expertise, obtaining such annotations can be costly or impractical.
FineSurE is designed to use human-provided keyfacts when available.
Alternatively, both keyfact extraction and alignment are automatically performed using the LLM with task-specific prompts, as shown in Appendix~\ref{appendix:prompt_finesure}.

Suppose that after keyfact extraction, a document \( X \) contains \( \hat{n} \) keyfacts and the summary \( Y \) contains \( \tilde{n} \) sentences.
During the keyfact alignment process, each keyfact may be aligned with one or more summary sentences, or may remain unaligned.
Let \( \hat{n}_{*} \) denote the number of keyfacts that can be aligned with at least one summary sentence, and \( \tilde{n}_{*} \) denote the number of summary sentences that can be aligned with at least one keyfact.
The completeness and conciseness scores are defined as
$S_{\text{com}}(\tilde{Y}) = S(\tilde{Y} | X, D = \text{com}) = \hat{n}_{*}/\hat{n}$ and
$S_{\text{con}}(\tilde{Y}) = S(\tilde{Y} | X, D = \text{con}) = \tilde{n}_{*}/\tilde{n}$, respectively.
These two fractions are intuitive proxy for summary quality, because completeness is measured as the proportion of document keyfacts that recoverable from the summary, indicating how much of the essential source content is preserved.
Conversely, conciseness is measured as the proportion of summary sentences that are aligned with keyfacts, capturing how much of the summary is semantically meaningful rather than redundant or off-topic. 
Together, they reflect a balance between including sufficient information and avoiding unnecessary verbosity.
Although completeness and conciseness are central to the utility of a summary, faithfulness, {\em i.e.}, whether the summary is factually consistent with the document constitutes a complementary but distinct dimension and is not directly assessed by these alignment-based scores.

\section{Prompt for FineSurE Calculation}\label{appendix:prompt_finesure}
Prompt for keyfact extraction:

{\small \textit{
You will be provided with a summary. Your task is to decompose
the summary into a set of "key facts". A "key fact" is a single
fact written as briefly and clearly as possible, encompassing at
most 2-3 entities.\\
Here are nine examples of key facts to illustrate the desired
level of granularity: \\
\textasteriskcentered\ Kevin Carr set off on his journey from Haytor. \\
\textasteriskcentered\ Kevin Carr set off on his journey from Dartmoor. \\
\textasteriskcentered\ Kevin Carr set off on his journey in July 2013. \\
\textasteriskcentered\ Kevin Carr is less than 24 hours away from completing his trip. \\
\textasteriskcentered\ Kevin Carr ran around the world unsupported. \\
\textasteriskcentered\ Kevin Carr ran with his tent. \\
\textasteriskcentered\ Kevin Carr is set to break the previous record. \\
\textasteriskcentered\ Kevin Carr is set to break the record by 24 hours. \\
\textasteriskcentered\ The previous record was held by an Australian. \\
Instruction: \\
First, read the summary carefully. \\
Second, decompose the summary into (at most 16) key facts. \\
Provide your answer in JSON format. The answer should be a
dictionary with the key "key facts" containing the key facts as a
list: \\
\{ "key facts": ["first key fact", "second key fact", "third key fact"], Summary: \textcolor{blue}{[summary]} \}
}}

Prompt for keyfact alignment:

{\small \textit{
You will receive a summary and a set of key facts for the same transcript. Your task is to assess if each key fact is inferred from the summary. \\
Instruction:
First, compare each key fact with the summary.
Second, check if the key fact is inferred from the summary and then response "Yes" or "No" for each key fact. If "Yes", specify the line number(s) of the summary sentence(s) relevant to each key fact. \\
Provide your answer in JSON format. The answer should be a list of dictionaries whose keys are "key fact", "response", and "line number":
[{"key fact": "first key fact", "response": "Yes", "line number": [1]}, {"key fact": "second key fact", "response": "No", "line number": []}, {"key fact": "third key fact", "response": "Yes", "line number": [1, 2, 3]}]\\
Summary:
\textcolor{blue}{[summary]}
\textcolor{blue}{[N]} key facts:
\textcolor{blue}{[key-facts]}
}}

\textcolor{blue}{Text in blue} indicates the part of the input that is fed into the prompt.

\section{Prompt for Control Generation}\label{appendix_control_prompt}
Prompt to control model generation prioritizing completeness over conciseness:

{\small \textit{Below is an instruction that describes a task. Write a response that appropriately completes the request. 
Instruction: Please summarize the input document, prioritizing completeness over conciseness. Return the summary in the following JSON format: {"Summary": "answer"} 
Input:\textcolor{blue}{[document]}
Response:}}

Prompt to control model generation prioritizing conciseness over completeness:

{\small \textit{Below is an instruction that describes a task. Write a response that appropriately completes the request. 
Instruction: Please summarize the input document, prioritizing conciseness over completeness. Return the summary in the following JSON format: {"Summary": "answer"} 
Input:\textcolor{blue}{[document]}
Response:}}

Prompt to control model generation balance completeness and conciseness:

{\small \textit{Below is an instruction that describes a task. Write a response that appropriately completes the request. 
Instruction: Please summarize the input document, balancing completeness with conciseness. Return the summary in the following JSON format: {"Summary": "answer"} 
Input:\textcolor{blue}{[document]}
Response:}}

Sometimes, the model may fail to provide the correct JSON format, making it difficult to extract the intended answer.
In such cases, it is often necessary to query the model multiple times to obtain a valid JSON output.
\textcolor{blue}{Text in blue} indicates the part of the input that is fed into the prompt.

\section{Model Hyperparameters}\label{appendix:hyperparams}
For training, we selected $K=15$, and used a batch size of 4, trained for 3 epochs. We set the hyperparameters as $\lambda = 0.5$, $\phi = 0.1$, $\gamma = 1$, and $\beta = 1$. For LoRA Training, we used lora\_alpha = 16, lora\_dropout  = 0, and target\_modules= \{["q\_proj", "k\_proj", "v\_proj", "o\_proj", "gate\_proj", "up\_proj", "down\_proj"]\}. For the model generation, we used nucleus (top-p) sampling {"generation\_config": GenerationConfig {"do\_sample": true, "temperature": 0.6, "top\_p": 0.9}}.

\section{Trade-off verification}\label{appendix:tradeoff_verification}

To verify that completeness and conciseness exhibit a clear trade-off relationship.
We used the FeedSum dataset, which contains 13,348 document–summary pairs along with their completeness and conciseness scores.
These scores were computed by G-Eval+ and FineSurE.
We observe that 67\% of the data points fall outside the extreme cases (1,1) and (0,0), suggesting meaningful variation between the two dimensions.

For these non-extreme points, we wanted to examine how one score behaves when the other becomes high.
We filtered samples where either completeness or conciseness is above a certain threshold, and then computed the Spearman correlation between the two metrics.

From Table~\ref{tab:tradeoff_spearman}, we can clearly see that as one dimension approaches a higher score, the Spearman correlation becomes more negative, indicating a stronger trade-off.
For G-Eval+ setting, the spearman correlation drops from –0.19 $\to$ –0.95.
For FineSurE, it drops from –0.04 $\to$ –0.85.

In addition to the Spearman correlation analysis presented above, we further performed a Pareto frontier analysis to make this relationship explicit.

Based on this analysis, we obtained the following results: using Completeness and Conciseness scores from G-Eval+, the Pareto-optimal region lies in Completeness $\in $ [0.800, 1.000] and Conciseness $\in $ [0.800, 1.000], with an average trade-off slope of –1.000, meaning that for every 1-unit increase in Completeness, Conciseness decreases by approximately 1.000 units.
Using FineSurE scores, the Pareto-optimal region lies in Completeness $\in $ [0.938, 1.000] and Conciseness $\in $ [0.889, 1.000], with an average trade-off slope of –1.778, meaning that for every 1-unit increase in Completeness, Conciseness decreases by roughly 1.778 units.
Both analyses clearly reveal a negative trade-off. Because this trade-off is substantial and well-defined, the problem addressed in our paper is substantiated and deemed of research value.


\begin{table}[t]
\centering
\small
\scalebox{0.8}{
\begin{tabular}{cccc}
\toprule
Threshold & Valid Docs & Spearman Mean & Spearman Std \\
\midrule
\multicolumn{4}{c}{\textbf{G-Eval+}} \\
\midrule
0.5 & 10116 & -0.19$\pm$0.50 & 0.50 \\
0.6 & 10116 & -0.19$\pm$0.50 & 0.50 \\
0.7 & 9427  & -0.41$\pm$0.40 & 0.40 \\
0.8 & 9427  & -0.41$\pm$0.40 & 0.40 \\
0.9 & 7809  & -0.95$\pm$0.08 & 0.08 \\
\midrule
\multicolumn{4}{c}{\textbf{FineSurE}} \\
\midrule
0.5 & 8515  & -0.04$\pm$0.47 & 0.47 \\
0.6 & 7335  & -0.27$\pm$0.51 & 0.51 \\
0.7 & 5633  & -0.48$\pm$0.51 & 0.51 \\
0.8 & 4196  & -0.60$\pm$0.47 & 0.47 \\
0.9 & 1916  & -0.85$\pm$0.13 & 0.13 \\
\bottomrule
\end{tabular}}
\caption{Spearman correlation between completeness and conciseness at different thresholds for non-extreme points in the FeedSum dataset (13,348 document-summary pairs). The mean and standard deviation are reported for each threshold.}
\label{tab:tradeoff_spearman}
\end{table}

\begin{table}[t]
\centering
\small
\scalebox{0.7}{
\begin{tabular}{lcccc}
\toprule
Model & Pear Med & Pear IQR & Ked Med & Ked IQR \\
\midrule
LLaMA       & 0.034  & (-0.244, 0.291) & 0.013  & (-0.244, 0.166) \\
SummLLaMA   & -0.037 & (-0.346, 0.304) & 0.035  & (-0.182, 0.297) \\
LLaMA*      & 0.195  & (-0.146, 0.459) & 0.114  & (-0.164, 0.295) \\
Qwen        & 0.027  & (-0.299, 0.305) & -0.034 & (-0.262, 0.210) \\
Qwen*       & 0.236  & (-0.021, 0.488) & 0.175  & (-0.024, 0.349) \\
Mistral     & 0.003  & (-0.240, 0.221) & -0.018 & (-0.177, 0.138) \\
Mistral*    & 0.109  & (-0.211, 0.396) & 0.066  & (-0.148, 0.304) \\
\bottomrule
\end{tabular}}
\caption{Extended results including both Pearson and Kendall correlation metrics. Pear and Ked indicates Pearson and Kendall correlation metrics, respectively. Med indicates Medians. 
* indicates our fine-tuned version. Fine-tuned models consistently outperform their respective baselines across all three model families (LLaMA, Qwen, and Mistral). For example, LLaMA* achieves a Pearson correlation of 0.195 versus 0.034 for LLaMA and –0.037 for SummLLaMA; for Kendall-tau, LLaMA* obtains 0.114 versus 0.013 for LLaMA and 0.035 for SummLLaMA. The best performance is achieved by our fine-tuned Qwen*. These additional findings further strengthen our conclusions.}
\label{tab:pearson_kendall}
\end{table}




\clearpage

\begin{table*}[th]
\hspace{1mm}
\centering
\begin{tabular}{llll}
\hline
Dataset        & Domain        & Training       & Test   \\ \hline
\multirow{3}{*}{FeedSum} & lifestyle       & \multirow{3}{*}{12000} & \multirow{3}{*}{600} \\
   & news        &  &  \\
   & daily life conversation &  &  \\
\cdashline{1-4}
OpoSum     & product reviews   & -  & 200        \\
MeQSum     & medical   & -  & 200        \\ \hline
\end{tabular}
\caption{Dataset summary.}
\label{tab:dataset_summary}
\end{table*}

\begin{figure*}[th]
    \centering
    \includegraphics[width=\columnwidth]{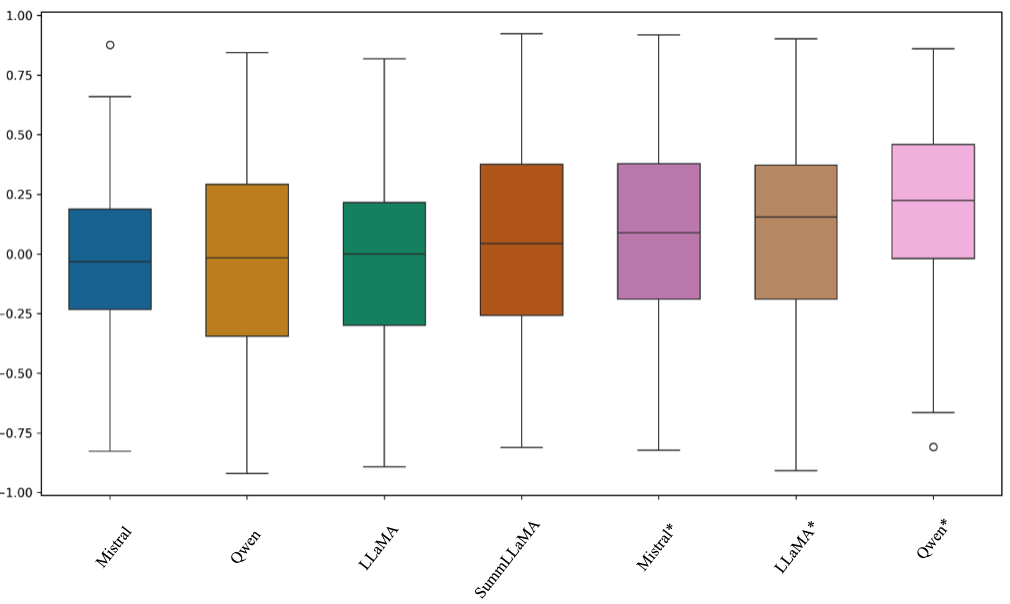}
    \caption{
    Distribution of Spearman correlations between model likelihood and model-based scores across different models. Models are sorted from left to right by median correlation value, from lowest to highest.
    }
    \label{fig:ranking_distribution}
\end{figure*}

\begin{table*}[th]
\centering
\scalebox{0.8}{
\begin{tabular}{llcccccc}
\toprule
\textbf{Model} & \textbf{Criteria} & $\mathbf{S_{com}}$        & $\mathbf{S_{con}}$        & $\mathbf{HM}(\tilde{Y})$   & $\mathbf{R}(\tilde{Y})$    & \textbf{Succ rate}  & \textbf{Prop. of 1} \\
\midrule
LLaMA            & Com$_{\uparrow}$  & 0.73 $\pm$ 0.24 & 0.52 $\pm$ 0.24 & 0.57 $\pm$ 0.18 & 0.38 $\pm$ 0.56 & 0.77 & 0.08 \\
LLaMA            & Con$_{\uparrow}$ & 0.57 $\pm$ 0.23 & 0.76 $\pm$ 0.26 & 0.60 $\pm$ 0.17 & 0.31 $\pm$ 0.60 & 0.65 & 0.22 \\
LLaMA            & Bal  & 0.70 $\pm$ 0.26 & 0.69 $\pm$ 0.27 & 0.66 $\pm$ 0.22 & 0.03 $\pm$ 0.51 & 0.84 & - \\
\cdashline{1-8}
SummLLaMA        & Com$_{\uparrow}$  & 0.63 $\pm$ 0.21 & 0.73 $\pm$ 0.28 & 0.62 $\pm$ 0.17 & -0.11 $\pm$ 0.66 & 0.65 & 0.24 \\
SummLLaMA        & Con$_{\uparrow}$ & 0.46 $\pm$ 0.18 & 0.99 $\pm$ 0.07 & 0.60 $\pm$ 0.17 & 0.86 $\pm$ 0.47  & 0.62 & 0.26 \\
SummLLaMA        & Bal  & 0.68 $\pm$ 0.27 & 0.97 $\pm$ 0.12 & 0.76 $\pm$ 0.21 & -0.44 $\pm$ 0.49 & 0.89 & - \\
\cdashline{1-8}
LLaMA*(MS)       & Com$_{\uparrow}$  & 0.75 $\pm$ 0.23 & 0.54 $\pm$ 0.21 & 0.60 $\pm$ 0.17 & 0.35 $\pm$ 0.51 & 0.78 & 0.07 \\
LLaMA*(MS)       & Con$_{\uparrow}$ & 0.56 $\pm$ 0.23 & 0.76 $\pm$ 0.26 & 0.60 $\pm$ 0.17 & 0.33 $\pm$ 0.60 & 0.66 & 0.23 \\
LLaMA*(MS)       & Bal  & 0.71 $\pm$ 0.25 & 0.70 $\pm$ 0.26 & 0.67 $\pm$ 0.22 & 0.02 $\pm$ 0.50 & 0.86 & - \\
\cdashline{1-8}
LLaMA*(CO)       & Com$_{\uparrow}$  & 0.74 $\pm$ 0.23 & 0.53 $\pm$ 0.23 & 0.58 $\pm$ 0.17 & 0.36 $\pm$ 0.56 & 0.78 & 0.10 \\
LLaMA*(CO)       & Con$_{\uparrow}$ & 0.56 $\pm$ 0.26 & 0.77 $\pm$ 0.26 & 0.60 $\pm$ 0.18 & 0.33 $\pm$ 0.59 & 0.65 & 0.24 \\
LLaMA*(CO)       & Bal  & 0.72 $\pm$ 0.25 & 0.70 $\pm$ 0.27 & 0.67 $\pm$ 0.22 & 0.04 $\pm$ 0.52 & 0.88 & - \\
\cdashline{1-8}
LLaMA*(MR+MS)    & Com$_{\uparrow}$  & 0.71 $\pm$ 0.22 & 0.63 $\pm$ 0.25 & 0.62 $\pm$ 0.16 & 0.15 $\pm$ 0.57 & 0.70 & 0.19 \\
LLaMA*(MR+MS)    & Con$_{\uparrow}$ & 0.53 $\pm$ 0.20 & 0.84 $\pm$ 0.24 & 0.63 $\pm$ 0.17 & 0.57 $\pm$ 0.52 & 0.57 & 0.31 \\
LLaMA*(MR+MS)    & Bal  & 0.73 $\pm$ 0.25 & 0.84 $\pm$ 0.24 & 0.75 $\pm$ 0.22 & -0.14 $\pm$ 0.47 & 0.89 & - \\
\cdashline{1-8}
LLaMA*(MR+CO)    & Com$_{\uparrow}$  & 0.74 $\pm$ 0.21 & 0.60 $\pm$ 0.24 & 0.61 $\pm$ 0.16 & 0.25 $\pm$ 0.53 & 0.71 & 0.18 \\
LLaMA*(MR+CO)    & Con$_{\uparrow}$ & 0.53 $\pm$ 0.19 & 0.88 $\pm$ 0.22 & 0.62 $\pm$ 0.16 & 0.54 $\pm$ 0.51 & 0.59 & 0.29 \\
LLaMA*(MR+CO)    & Bal  & 0.74 $\pm$ 0.25 & 0.80 $\pm$ 0.25 & 0.73 $\pm$ 0.22 & -0.09 $\pm$ 0.47 & 0.90 & - \\
\cdashline{1-8}
LLaMA*(MS+CO)    & Com$_{\uparrow}$  & 0.75 $\pm$ 0.23 & 0.53 $\pm$ 0.22 & 0.58 $\pm$ 0.18 & 0.38 $\pm$ 0.52 & 0.80 & 0.10 \\
LLaMA*(MS+CO)    & Con$_{\uparrow}$ & 0.55 $\pm$ 0.22 & 0.77 $\pm$ 0.26 & 0.60 $\pm$ 0.18 & 0.36 $\pm$ 0.59 & 0.63 & 0.24 \\
LLaMA*(MS+CO)    & Bal  & 0.72 $\pm$ 0.25 & 0.72 $\pm$ 0.26 & 0.69 $\pm$ 0.21 & 0.01 $\pm$ 0.51 & 0.88 & - \\
\cdashline{1-8}
LLaMA*(MR+MS+CO) & Com$_{\uparrow}$  & 0.72 $\pm$ 0.22 & 0.64 $\pm$ 0.25 & 0.63 $\pm$ 0.17 & 0.15 $\pm$ 0.55 & 0.71 & 0.19 \\
LLaMA*(MR+MS+CO) & Con$_{\uparrow}$ & 0.52 $\pm$ 0.19 & 0.92 $\pm$ 0.18 & 0.63 $\pm$ 0.16 & 0.62 $\pm$ 0.53 & 0.60 & 0.31 \\
LLaMA*(MR+MS+CO) & Bal  & 0.75 $\pm$ 0.25 & 0.85 $\pm$ 0.23 & 0.76 $\pm$ 0.21 & -0.15 $\pm$ 0.48 & 0.92 & - \\
\hline
\end{tabular}}
\caption{Ablation study on different loss components in LLaMA fine-tuning. We compare model performance across variants of LLaMA fine-tuned with different combinations of loss components using FineSurE. The reported values are averages across all FeedSum cases, with standard deviations in parentheses. $*$ indicates the base LLaMA model fine-tuned with our full objective. Com$_{\uparrow}$ prioritizes completeness, Con$_{\uparrow}$ prioritizes conciseness, while Bal aims to strike a balance between the two objectives. Succ rate denotes the percentage of examples where either $S_\text{com}$ or $S_\text{con}$ is non-zero. Proportion of 1 refers to cases where both $S_\text{com}$ and $S_\text{con}$ are exactly 1. These are considered uncontrollable in the Com${\uparrow}$ and Con${\uparrow}$ settings and are excluded when calculating $S_\text{com}$, $S_\text{con}$, $\text{HM}(\tilde{Y})$, and $\text{R}(\tilde{Y})$. Under the Bal objective, however, such cases are regarded as ideal and thus retained in evaluation (denoted as “–”). The MR-only variant is excluded here due to frequent repetition issues it introduces during generation.}
\label{tab:ablation_finesure}
\end{table*}



\begin{table*}[th]
\hspace{1mm}
\centering
\scalebox{0.6}{
\begin{tabular}{clcccccccc}
\hline
\multicolumn{1}{l}{\textbf{Model}} & \textbf{Category} & \textbf{Consistency} & \textbf{Coherence} & \textbf{Fluency} & \textbf{Relevance} & \textbf{ROUGE-1} & \textbf{ROUGE-2} & \textbf{ROUGE-3} & \textbf{BERTScore} \\ \hline
\multirow{3}{*}{LLaMA}             & Com$_{\uparrow}$              & $3.50 \pm 1.19$      & $2.11 \pm 1.31$    & $0.84 \pm 0.78$  & $3.06 \pm 1.25$    & $0.23 \pm 0.13$  & $0.08 \pm 0.08$  & $0.04 \pm 0.06$  & $0.30 \pm 0.15$    \\
                                   & Con$_{\uparrow}$              & $2.87 \pm 0.88$      & $1.58 \pm 1.28$    & $0.44 \pm 0.55$  & $2.49 \pm 1.25$    & $0.28 \pm 0.14$  & $0.09 \pm 0.09$  & $0.04 \pm 0.06$  & $0.29 \pm 0.17$    \\
                                   & Bal               & $3.29 \pm 1.30$      & $1.91 \pm 1.32$    & $0.63 \pm 0.73$  & $2.76 \pm 1.25$    & $0.27 \pm 0.14$  & $0.09 \pm 0.09$  & $0.04 \pm 0.06$  & $0.30 \pm 0.15$    \\ \cdashline{1-10}
\multirow{3}{*}{SummLLaMA}         & Com$_{\uparrow}$              & $3.53 \pm 1.21$      & $2.02 \pm 1.24$    & $0.72 \pm 0.74$  & $3.00 \pm 1.26$    & $0.23 \pm 0.13$  & $0.08 \pm 0.07$  & $0.04 \pm 0.05$  & $0.31 \pm 0.13$    \\
                                   & Con$_{\uparrow}$              & $2.48 \pm 0.79$      & $1.30 \pm 1.08$    & $0.33 \pm 0.44$  & $2.22 \pm 1.25$    & $0.28 \pm 0.14$  & $0.09 \pm 0.10$  & $0.04 \pm 0.07$  & $0.27 \pm 0.17$    \\
                                   & Bal               & $3.34 \pm 1.15$      & $1.67 \pm 1.20$    & $0.45 \pm 0.55$  & $2.58 \pm 1.22$    & $0.27 \pm 0.14$  & $0.10 \pm 0.09$  & $0.04 \pm 0.06$  & $0.29 \pm 0.15$    \\ \cdashline{1-10}
\multirow{3}{*}{LLaMA*}            & Com$_{\uparrow}$              & $3.48 \pm 1.24$      & $2.08 \pm 1.28$    & $0.85 \pm 0.78$  & $3.19 \pm 1.27$    & $0.23 \pm 0.12$  & $0.08 \pm 0.07$  & $0.04 \pm 0.05$  & $0.31 \pm 0.14$    \\
                                   & Con$_{\uparrow}$              & $2.89 \pm 0.83$      & $1.63 \pm 1.21$    & $0.49 \pm 0.53$  & $2.61 \pm 1.27$    & $0.28 \pm 0.14$  & $0.09 \pm 0.09$  & $0.04 \pm 0.06$  & $0.28 \pm 0.16$    \\
                                   & Bal               & $3.38 \pm 1.17$      & $1.88 \pm 1.22$    & $0.65 \pm 0.72$  & $2.92 \pm 1.25$    & $0.27 \pm 0.13$  & $0.10 \pm 0.08$  & $0.04 \pm 0.06$  & $0.30 \pm 0.15$    \\ \hline
\end{tabular}}
\caption{Detailed comparison of model performance across multiple quality dimensions. We evaluate consistency, coherence, fluency, and relevance using G-Eval, reporting averages over all cases with standard deviations. ROUGE and BERTScore are also included, computed based on human reference summaries. $*$ denotes foundation models fine-tuned using our method. These metrics provide a more comprehensive assessment of summary quality beyond traditional reference-based measures.}
\label{tab:g_eval}
\end{table*}

\begin{table*}[th]
\vspace{1mm}
\centering
\scalebox{0.8}{
\label{tab:model_comparison}
\begin{tabular}{llcccccc}
\toprule
\textbf{Model} & \textbf{Criteria} & $\mathbf{S_{com}}$        & $\mathbf{S_{con}}$        & $\mathbf{HM}(\tilde{Y})$   & $\mathbf{R}(\tilde{Y})$    & \textbf{Succ rate}  & \textbf{Prop. of 1} \\
\midrule

\multirow{3}{*}{LLaMA}      & Com$_{\uparrow}$  & 0.73 $\pm$ 0.24                            & 0.52 $\pm$ 0.24                            & 0.57 $\pm$ 0.18                 & 0.38 $\pm$ 0.56                    & 0.77               & 0.08                \\
                            & Con$_{\uparrow}$  & 0.57 $\pm$ 0.23                            & 0.76 $\pm$ 0.26                            & 0.60 $\pm$ 0.17                 & 0.31 $\pm$ 0.6                     & 0.65               & 0.22                \\
                            & Bal               & 0.70 $\pm$ 0.26                            & 0.69 $\pm$ 0.27                            & 0.66 $\pm$ 0.22                 & 0.03 $\pm$ 0.51                    & 0.84               & -                   \\
\cdashline{1-8}
\multirow{3}{*}{SummLLaMA}  & Com$_{\uparrow}$  & 0.63 $\pm$ 0.21                            & 0.73 $\pm$ 0.28                            & 0.62 $\pm$ 0.17                 & -0.11 $\pm$ 0.66                   & 0.65               & 0.24                \\
                            & Con$_{\uparrow}$  & 0.46 $\pm$ 0.18                            & 0.99 $\pm$ 0.07                            & 0.60 $\pm$ 0.17                 & 0.86 $\pm$ 0.47                    & 0.62               & 0.26                \\
                            & Bal               & 0.68 $\pm$ 0.27                            & 0.97 $\pm$ 0.12                            & 0.76 $\pm$ 0.21                 & -0.44 $\pm$ 0.49                   & 0.89               & -                   \\
\cdashline{1-8}
\multirow{3}{*}{LLaMA*}     & Com$_{\uparrow}$  & 0.72 $\pm$ 0.22                            & 0.64 $\pm$ 0.25                            & 0.63 $\pm$ 0.17                 & 0.15 $\pm$ 0.55                    & 0.71               & 0.19                \\
                            & Con$_{\uparrow}$  & 0.52 $\pm$ 0.19                            & 0.92 $\pm$ 0.18                            & 0.63 $\pm$ 0.16                 & 0.62 $\pm$ 0.53                    & 0.60               & 0.31                \\
                            & Bal               & 0.75 $\pm$ 0.25                            & 0.85 $\pm$ 0.23                            & 0.76 $\pm$ 0.21                 & -0.15 $\pm$ 0.48                   & 0.92               & -                   \\
\cdashline{1-8}
\multirow{3}{*}{Qwen}       & Com$_{\uparrow}$  & 0.74 $\pm$ 0.23                            & 0.52 $\pm$ 0.21                            & 0.57 $\pm$ 0.17                 & 0.40 $\pm$ 0.52                    & 0.81               & 0.08                \\
                            & Con$_{\uparrow}$  & 0.64 $\pm$ 0.24                            & 0.64 $\pm$ 0.24                            & 0.60 $\pm$ 0.17                 & -0.01 $\pm$ 0.53                   & 0.71               & 0.16                \\
                            & Bal               & 0.72 $\pm$ 0.25                            & 0.64 $\pm$ 0.25                            & 0.65 $\pm$ 0.21                 & 0.13 $\pm$ 0.49                    & 0.88               & -                   \\
\cdashline{1-8}
\multirow{3}{*}{Qwen*}      & Com$_{\uparrow}$  & 0.66 $\pm$ 0.23                            & 0.63 $\pm$ 0.25                            & 0.60 $\pm$ 0.16                 & 0.05 $\pm$ 0.6                     & 0.70               & 0.19                \\
                            & Con$_{\uparrow}$  & 0.54 $\pm$ 0.21                            & 0.78 $\pm$ 0.25                            & 0.60 $\pm$ 0.18                 & 0.39 $\pm$ 0.53                    & 0.63               & 0.24                \\
                            & Bal               & 0.71 $\pm$ 0.26                            & 0.75 $\pm$ 0.26                            & 0.70 $\pm$ 0.22                 & -0.06 $\pm$ 0.48                   & 0.87               & -                   \\
\cdashline{1-8}
\multirow{3}{*}{Mistral}    & Com$_{\uparrow}$  & 0.73 $\pm$ 0.23                            & 0.53 $\pm$ 0.22                            & 0.58 $\pm$ 0.18                 & 0.36 $\pm$ 0.53                    & 0.79               & 0.08                \\
                            & Con$_{\uparrow}$  & 0.63 $\pm$ 0.24                            & 0.63 $\pm$ 0.26                            & 0.59 $\pm$ 0.17                 & -0.01 $\pm$ 0.57                   & 0.70               & 0.18                \\
                            & Bal               & 0.73 $\pm$ 0.25                            & 0.64 $\pm$ 0.26                            & 0.64 $\pm$ 0.22                 & 0.15 $\pm$ 0.51                    & 0.88               & -                   \\
\cdashline{1-8}
\multirow{3}{*}{Mistral*}   & Com$_{\uparrow}$  & 0.67 $\pm$ 0.24                            & 0.68 $\pm$ 0.27                            & 0.62 $\pm$ 0.18                 & 0.02 $\pm$ 0.64                    & 0.63               & 0.20                \\
                            & Con$_{\uparrow}$  & 0.56 $\pm$ 0.22                            & 0.75 $\pm$ 0.27                            & 0.59 $\pm$ 0.17                 & 0.30 $\pm$ 0.6                     & 0.63               & 0.24                \\
                            & Bal               & 0.71 $\pm$ 0.26                            & 0.74 $\pm$ 0.27                            & 0.69 $\pm$ 0.23                 & -0.03 $\pm$ 0.5                    & 0.87               & -                   \\
\cdashline{1-8}
\multirow{3}{*}{GPT4omini}  & Com$_{\uparrow}$  & 0.83 $\pm$ 0.21                            & 0.48 $\pm$ 0.2                             & 0.58 $\pm$ 0.18                 & 0.6 $\pm$ 0.51                     & 0.84               & 0.04                \\
                            & Con$_{\uparrow}$  & 0.67 $\pm$ 0.23                            & 0.64 $\pm$ 0.25                            & 0.61 $\pm$ 0.17                 & -0.06 $\pm$ 0.53                   & 0.69               & 0.2                 \\
                            & Bal               & 0.78 $\pm$ 0.23                            & 0.62 $\pm$ 0.25                            & 0.66 $\pm$ 0.21                 & 0.29 $\pm$ 0.51                    & 0.89               & -                   \\
\cdashline{1-8}
\multirow{3}{*}{GPT4o}      & Com$_{\uparrow}$  & 0.82 $\pm$ 0.21                            & 0.47 $\pm$ 0.2                             & 0.57 $\pm$ 0.18                 & 0.62 $\pm$ 0.49                    & 0.85               & 0.06                \\
                            & Con$_{\uparrow}$  & 0.66 $\pm$ 0.23                            & 0.61 $\pm$ 0.24                            & 0.59 $\pm$ 0.17                 & -0.08 $\pm$ 0.53                   & 0.69               & 0.19                \\
                            & Bal               & 0.78 $\pm$ 0.23                            & 0.62 $\pm$ 0.26                            & 0.66 $\pm$ 0.22                 & 0.27 $\pm$ 0.52                    & 0.89               & -                   \\
\cdashline{1-8}
\multirow{3}{*}{Keyfact\_GPT4o }               & Com$_{\uparrow}$        & $0.72 \pm 0.24$ & $0.56 \pm 0.22$ & $0.59 \pm 0.18$      & $0.27 \pm 0.51$     & $0.78$             & $0.12$     \\
                                              & Con$_{\uparrow}$       & $0.56 \pm 0.23$ & $0.71 \pm 0.26$ & $0.58 \pm 0.18$      & $0.26 \pm 0.56$     & $0.62$             & $0.24 $     \\
                                              & Bal           & $0.70 \pm 0.27$ & $0.74 \pm 0.26$ & $0.69 \pm 0.23$      & $-0.06 \pm 0.47$    & $0.87$             & -    \\ 
\cdashline{1-8}
\multirow{3}{*}{Keyfact\_GPT4o* }               & Com$_{\uparrow}$ & $0.73 \pm 0.21$ & $0.63 \pm 0.27$ & $0.63 \pm 0.17$ & $0.14 \pm 0.50$  & $0.71$ & $0.20$  \\
                                                                      & Con$_{\uparrow}$ & $0.52 \pm 0.26$ & $0.93 \pm 0.26$ & $0.63 \pm 0.18$ & $0.65 \pm 0.50$  & $0.60$ & $0.32 $ \\
                                                                      & Bal              & $0.74 \pm 0.23$ & $0.84 \pm 0.21$ & $0.75 \pm 0.25$ & $-0.16 \pm 0.51$ & $0.91$ & -       \\
\cdashline{1-8}
\multirow{3}{*}{GPT4-turbo} & Com$_{\uparrow}$  & 0.88 $\pm$ 0.17                            & 0.61 $\pm$ 0.2                             & 0.69 $\pm$ 0.15                 & 0.42 $\pm$ 0.45                    & 0.77               & 0.18                \\
                            & Con$_{\uparrow}$  & 0.69 $\pm$ 0.2                             & 0.82 $\pm$ 0.21                            & 0.71 $\pm$ 0.14                 & 0.18 $\pm$ 0.49                    & 0.54               & 0.42                \\
                            & Bal               & 0.86 $\pm$ 0.19                            & 0.81 $\pm$ 0.22                            & 0.81 $\pm$ 0.18                 & 0.07 $\pm$ 0.38                    & 0.95               & -                   \\
\cdashline{1-8}
\multirow{3}{*}{Gemini}     & Com$_{\uparrow}$  & 0.86 $\pm$ 0.2                             & 0.45 $\pm$ 0.19                            & 0.57 $\pm$ 0.18                 & 0.71 $\pm$ 0.49                    & 0.86               & 0.03                \\
                            & Con$_{\uparrow}$  & 0.61 $\pm$ 0.24                            & 0.66 $\pm$ 0.25                            & 0.59 $\pm$ 0.16                 & 0.08 $\pm$ 0.58                    & 0.67               & 0.22                \\
                            & Bal               & 0.76 $\pm$ 0.24                            & 0.66 $\pm$ 0.26                            & 0.67 $\pm$ 0.21                 & 0.16 $\pm$ 0.49                    & 0.88               & -                   \\ \hline

\end{tabular}}
\caption{Detailed comparison of model performance using FineSurE. The value shows averages over all cases in FeedSum with standard deviations. $*$ indicates the foundation model fine-tuned using our method. Com$_{\uparrow}$ prioritizes completeness, Con$_{\uparrow}$ prioritizes conciseness, and Bal aims to balance both. \textit{Succ rate} denotes the proportion of cases where either $S_\text{com}$ or $S_\text{con}$ is non-zero. \textit{Proportion of 1} indicates the proportion of cases where both $S_\text{com}$ and $S_\text{con}$ equal 1. In the Com$_{\uparrow}$ and Con$_{\uparrow}$ settings, these are considered uncontrollable cases and are excluded when computing $S_\text{com}$, $S_\text{con}$, $\text{HM}(\tilde{Y})$, and $\text{R}(\tilde{Y})$. In contrast, under Bal, these cases are treated as ideal (i.e., perfect), so they are retained (denoted as “–”). 
Key\_GPT4o is a prompt-oriented baseline. In Key\_GPT4o, we first use GPT-4o to extract key facts, which are then incorporated into a subsequent prompt for the summarizer. Key\_GPT4o* is a single-pass variant, where we prompt the model to first generate key facts and then produce a summary conditioned on these key facts within a single prompt.  
It is worth noting that if the model used for model-based scoring shares the same architecture as the model being evaluated ({\em e.g.}, using FineSurE(GPT-4o) to evaluate GPT-4o), the resulting score may be biased. In such cases, model-based scoring tends to favor models with similar architectures \citep{wataoka2024self, xu2024pride}.}
\label{tab:model_full_comparison}
\end{table*}

\begin{table*}[th]
\vspace{1mm}
\centering
\scalebox{0.8}{
\begin{tabular}{llll}
\hline
Model                       & Criteria         & Faithfulness    & AvgFaith              \\ \hline
\multirow{3}{*}{LLaMA}      & Com$_{\uparrow}$ & 0.83 $\pm$ 0.26 & \multirow{3}{*}{0.82} \\
                            & Con$_{\uparrow}$ & 0.83 $\pm$ 0.33 &                       \\
                            & Bal              & 0.8 $\pm$ 0.32  &                       \\
\cdashline{1-4}
\multirow{3}{*}{SummLLaMA}  & Com$_{\uparrow}$ & 0.85 $\pm$ 0.32 & \multirow{3}{*}{0.85} \\
                            & Con$_{\uparrow}$ & 0.85 $\pm$ 0.35 &                       \\
                            & Bal              & 0.84 $\pm$ 0.36 &                       \\
\cdashline{1-4}
\multirow{3}{*}{LLaMA*}     & Com$_{\uparrow}$ & 0.85 $\pm$ 0.3  & \multirow{3}{*}{0.82} \\
                            & Con$_{\uparrow}$ & 0.8 $\pm$ 0.42  &                       \\
                            & Bal              & 0.8 $\pm$ 0.38  &                       \\
\cdashline{1-4}
\multirow{3}{*}{Qwen}       & Com$_{\uparrow}$ & 0.88 $\pm$ 0.2  & \multirow{3}{*}{0.85} \\
                            & Con$_{\uparrow}$ & 0.83 $\pm$ 0.29 &                       \\
                            & Bal              & 0.85 $\pm$ 0.25 &                       \\
\cdashline{1-4}
\multirow{3}{*}{Qwen*}      & Com$_{\uparrow}$ & 0.87 $\pm$ 0.31 & \multirow{3}{*}{0.85} \\
                            & Con$_{\uparrow}$ & 0.83 $\pm$ 0.37 &                       \\
                            & Bal              & 0.85 $\pm$ 0.32 &                       \\
\cdashline{1-4}
\multirow{3}{*}{Mistral}    & Com$_{\uparrow}$ & 0.91 $\pm$ 0.17 & \multirow{3}{*}{0.9}  \\
                            & Con$_{\uparrow}$ & 0.89 $\pm$ 0.24 &                       \\
                            & Bal              & 0.9 $\pm$ 0.22  &                       \\
\cdashline{1-4}
\multirow{3}{*}{Mistral*}   & Com$_{\uparrow}$ & 0.81 $\pm$ 0.4  & \multirow{3}{*}{0.81} \\
                            & Con$_{\uparrow}$ & 0.82 $\pm$ 0.41 &                       \\
                            & Bal              & 0.8 $\pm$ 0.41  &                       \\
\cdashline{1-4}
\multirow{3}{*}{GPT4omini}  & Com$_{\uparrow}$ & 0.95 $\pm$ 0.13 & \multirow{3}{*}{0.94} \\
                            & Con$_{\uparrow}$ & 0.93 $\pm$ 0.19 &                       \\
                            & Bal              & 0.93 $\pm$ 0.16 &                       \\
\cdashline{1-4}
\multirow{3}{*}{GPT4o}      & Com$_{\uparrow}$ & 0.94 $\pm$ 0.14 & \multirow{3}{*}{0.93} \\
                            & Con$_{\uparrow}$ & 0.91 $\pm$ 0.22 &                       \\
                            & Bal              & 0.95 $\pm$ 0.14 &                       \\
\cdashline{1-4}
\multirow{3}{*}{GPT4-turbo} & Com$_{\uparrow}$ & 0.95 $\pm$ 0.12 &                   \\
                            & Con$_{\uparrow}$ & 0.92 $\pm$ 0.19 &   0.93                    \\
                            & Bal              & 0.92 $\pm$ 0.2  &                       \\
\cdashline{1-4}
\multirow{3}{*}{Gemini}     & Com$_{\uparrow}$ & 0.96 $\pm$ 0.12 & \multirow{3}{*}{0.93} \\
                            & Con$_{\uparrow}$ & 0.89 $\pm$ 0.24 &                       \\
                            & Bal              & 0.93 $\pm$ 0.18 &                       \\ \cline{1-4}
\end{tabular}}
\caption{Faithfulness evaluation results using FineSurE across all models. The value shows averages over all cases in FeedSum with standard deviations. AvgFaith indicates the average of faithfulness across different criteria.
}
\label{tb:faithfulness}
\end{table*}


\begin{figure*}[th]
    \centering
    \includegraphics[width=400pt]{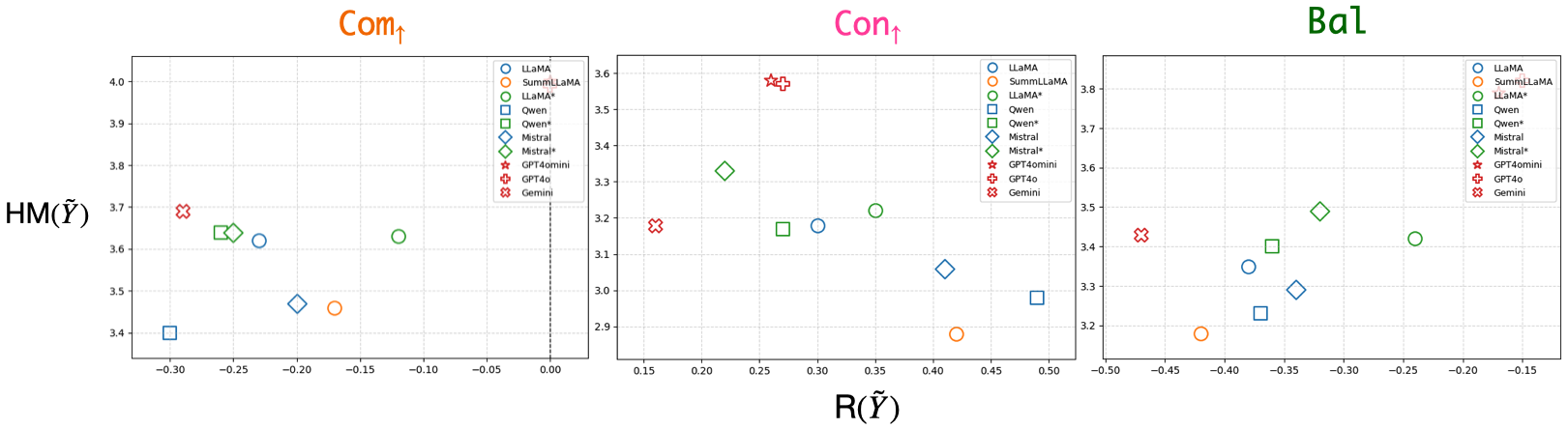}
    \caption{\small Scatter plot of model performance across different control settings using G-Eval+. Each point represents the mean of $\text{HM}(\tilde{Y})$ (y-axis) and $\text{R}(\tilde{Y})$ (x-axis) across all test cases in FeedSum test set. Models are grouped by color into four categories: Baseline models (blue), Our methods (green), SummLLaMA (orange), and Commercial models (red). The three panels show performance under different control priorities: Com$_{\uparrow}$ prioritizes completeness, Con$_{\uparrow}$  prioritizes conciseness, and Bal aims to balance both. The vertical dashed line at $\text{R}(\tilde{Y})$ = 0 represents the target equilibrium between completeness and conciseness, providing a reference for model controllability.}
    \label{fig:g-eval}
\end{figure*}

\begin{table*}[th]
\centering
\scalebox{0.7}{
\begin{tabular}{llllllll}
\hline
\textbf{Model}              & \textbf{Criteria} & $\mathbf{S_{com}}$ & $\mathbf{S_{con}}$ & $\mathbf{HM}(\tilde{Y})$     & $\mathbf{R}(\tilde{Y})$   & \textbf{Faithfulness} & \textbf{AvgFaith}     \\ \hline
\multirow{3}{*}{LLaMA}      & Com$_{\uparrow}$  & 3.51 $\pm$ 0.93        & 3.92 $\pm$ 0.59        & 3.62 $\pm$ 0.71 & -0.23 $\pm$ 0.37 & 0.88 $\pm$ 0.17       & \multirow{3}{*}{0.89} \\
                            & Con$_{\uparrow}$  & 2.79 $\pm$ 0.9         & 3.95 $\pm$ 0.67        & 3.18 $\pm$ 0.76 & 0.3 $\pm$ 0.43   & 0.91 $\pm$ 0.2        &                       \\
                            & Bal               & 3.06 $\pm$ 0.92        & 3.94 $\pm$ 0.65        & 3.35 $\pm$ 0.76 & -0.38 $\pm$ 0.39 & 0.88 $\pm$ 0.21       &                       \\
\cdashline{1-8}
\multirow{3}{*}{SummLLaMA}  & Com$_{\uparrow}$  & 3.43 $\pm$ 0.89        & 3.66 $\pm$ 0.64        & 3.46 $\pm$ 0.69 & -0.17 $\pm$ 0.36 & 0.94 $\pm$ 0.14       & \multirow{3}{*}{0.94} \\
                            & Con$_{\uparrow}$  & 2.4 $\pm$ 0.82         & 3.86 $\pm$ 0.71        & 2.88 $\pm$ 0.73 & 0.42 $\pm$ 0.4   & 0.94 $\pm$ 0.15       &                       \\
                            & Bal               & 2.84 $\pm$ 0.84        & 3.82 $\pm$ 0.65        & 3.18 $\pm$ 0.71 & -0.42 $\pm$ 0.37 & 0.95 $\pm$ 0.13       &                       \\
\cdashline{1-8}
\multirow{3}{*}{LLaMA*}     & Com$_{\uparrow}$  & 3.54 $\pm$ 0.89        & 3.87 $\pm$ 0.59        & 3.63 $\pm$ 0.67 & -0.12 $\pm$ 0.32 & 0.91 $\pm$ 0.21       & \multirow{3}{*}{0.91} \\
                            & Con$_{\uparrow}$  & 2.86 $\pm$ 0.86        & 3.9 $\pm$ 0.67         & 3.22 $\pm$ 0.71 & 0.35 $\pm$ 0.4   & 0.91 $\pm$ 0.27       &                       \\
                            & Bal               & 3.19 $\pm$ 0.87        & 3.89 $\pm$ 0.6         & 3.42 $\pm$ 0.69 & -0.24 $\pm$ 0.36 & 0.9 $\pm$ 0.25        &                       \\
\cdashline{1-8}
\multirow{3}{*}{Qwen}       & Com$_{\uparrow}$  & 3.1 $\pm$ 0.9          & 4 $\pm$ 0.62           & 3.4 $\pm$ 0.73  & -0.3 $\pm$ 0.39  & 0.9 $\pm$ 0.22        & \multirow{3}{*}{0.9}  \\
                            & Con$_{\uparrow}$  & 2.5 $\pm$ 0.77         & 3.93 $\pm$ 0.73        & 2.98 $\pm$ 0.69 & 0.49 $\pm$ 0.38  & 0.91 $\pm$ 0.24       &                       \\
                            & Bal               & 2.87 $\pm$ 0.89        & 3.95 $\pm$ 0.7         & 3.23 $\pm$ 0.76 & -0.37 $\pm$ 0.43 & 0.9 $\pm$ 0.24        &                       \\
\cdashline{1-8}
\multirow{3}{*}{Qwen*}      & Com$_{\uparrow}$  & 3.48 $\pm$ 0.9         & 3.99 $\pm$ 0.56        & 3.64 $\pm$ 0.7  & -0.26 $\pm$ 0.41 & 0.91 $\pm$ 0.14       & \multirow{3}{*}{0.91} \\
                            & Con$_{\uparrow}$  & 2.82 $\pm$ 0.87        & 3.89 $\pm$ 0.73        & 3.17 $\pm$ 0.74 & 0.27 $\pm$ 0.41  & 0.91 $\pm$ 0.18       &                       \\
                            & Bal               & 3.13 $\pm$ 0.94        & 3.95 $\pm$ 0.69        & 3.4 $\pm$ 0.77  & -0.36 $\pm$ 0.41 & 0.9 $\pm$ 0.16        &                       \\
\cdashline{1-8}
\multirow{3}{*}{Mistral}    & Com$_{\uparrow}$  & 3.47 $\pm$ 0.85        & 4 $\pm$ 0.55           & 3.64 $\pm$ 0.63 & -0.25 $\pm$ 0.32 & 0.92 $\pm$ 0.23       & \multirow{3}{*}{0.92} \\
                            & Con$_{\uparrow}$  & 3.02 $\pm$ 0.89        & 3.94 $\pm$ 0.68        & 3.33 $\pm$ 0.76 & 0.22 $\pm$ 0.42  & 0.92 $\pm$ 0.25       &                       \\
                            & Bal               & 3.24 $\pm$ 0.88        & 3.98 $\pm$ 0.63        & 3.49 $\pm$ 0.7  & -0.32 $\pm$ 0.36 & 0.92 $\pm$ 0.25       &                       \\
\cdashline{1-8}
\multirow{3}{*}{Mistral*}   & Com$_{\uparrow}$  & 3.26 $\pm$ 0.88        & 3.89 $\pm$ 0.66        & 3.47 $\pm$ 0.69 & -0.2 $\pm$ 0.32  & 0.91 $\pm$ 0.14       & \multirow{3}{*}{0.92} \\
                            & Con$_{\uparrow}$  & 2.66 $\pm$ 0.89        & 3.83 $\pm$ 0.78        & 3.06 $\pm$ 0.79 & 0.41 $\pm$ 0.36  & 0.92 $\pm$ 0.17       &                       \\
                            & Bal               & 2.93 $\pm$ 0.89        & 3.96 $\pm$ 0.66        & 3.29 $\pm$ 0.73 & -0.34 $\pm$ 0.36 & 0.93 $\pm$ 0.14       &                       \\
\cdashline{1-8}
\multirow{3}{*}{GPT4omini}  & Com$_{\uparrow}$  & 4.08 $\pm$ 0.77        & 4.03 $\pm$ 0.47        & 4 $\pm$ 0.54    & 0 $\pm$ 0.26     & 0.97 $\pm$ 0.09       & \multirow{3}{*}{0.98} \\
                            & Con$_{\uparrow}$  & 3.32 $\pm$ 0.97        & 4.13 $\pm$ 0.64        & 3.58 $\pm$ 0.77 & 0.26 $\pm$ 0.4   & 0.98 $\pm$ 0.08       &                       \\
                            & Bal               & 3.63 $\pm$ 0.92        & 4.16 $\pm$ 0.58        & 3.79 $\pm$ 0.71 & -0.17 $\pm$ 0.37 & 0.99 $\pm$ 0.07       &                       \\
\cdashline{1-8}
\multirow{3}{*}{GPT4o}      & Com$_{\uparrow}$  & 4.09 $\pm$ 0.78        & 4 $\pm$ 0.44           & 3.99 $\pm$ 0.57 & 0 $\pm$ 0.29     & 0.97 $\pm$ 0.09       & \multirow{3}{*}{0.98} \\
                            & Con$_{\uparrow}$  & 3.29 $\pm$ 0.91        & 4.14 $\pm$ 0.64        & 3.57 $\pm$ 0.74 & 0.27 $\pm$ 0.39  & 0.98 $\pm$ 0.1        &                       \\
                            & Bal               & 3.67 $\pm$ 0.88        & 4.15 $\pm$ 0.59        & 3.82 $\pm$ 0.69 & -0.15 $\pm$ 0.36 & 0.98 $\pm$ 0.07       &                       \\
\cdashline{1-8}
\multirow{3}{*}{GPT4-turbo} & Com$_{\uparrow}$  & 3.91 $\pm$ 0.93        & 4.02 $\pm$ 0.48        & 3.88 $\pm$ 0.67 & -0.06 $\pm$ 0.36 & 0.98 $\pm$ 0.08       & \multirow{3}{*}{0.98} \\
                            & Con$_{\uparrow}$  & 3.11 $\pm$ 0.99        & 4.05 $\pm$ 0.68        & 3.41 $\pm$ 0.85 & 0.33 $\pm$ 0.52  & 0.98 $\pm$ 0.09       &                       \\
                            & Bal               & 3.52 $\pm$ 0.96        & 4.08 $\pm$ 0.62        & 3.68 $\pm$ 0.76 & -0.19 $\pm$ 0.42 & 0.98 $\pm$ 0.09       &                       \\
\cdashline{1-8}
\multirow{3}{*}{Gemini}     & Com$_{\uparrow}$  & 3.7 $\pm$ 0.95         & 3.86 $\pm$ 0.52        & 3.69 $\pm$ 0.69 & -0.29 $\pm$ 0.39 & 0.95 $\pm$ 0.09       & \multirow{3}{*}{0.95} \\
                            & Con$_{\uparrow}$  & 2.82 $\pm$ 0.95        & 3.94 $\pm$ 0.69        & 3.18 $\pm$ 0.82 & 0.16 $\pm$ 0.46  & 0.95 $\pm$ 0.11       &                       \\
                            & Bal               & 3.18 $\pm$ 0.92        & 3.96 $\pm$ 0.65        & 3.43 $\pm$ 0.74 & -0.47 $\pm$ 0.41 & 0.95 $\pm$ 0.09       &                       \\ \cline{1-8}

\end{tabular}
}
\vspace{5pt}
\caption{Detailed comparison of model performance using G-Eval+. The value shows averages over all cases in FeedSum with standard deviations. $*$ denotes foundation models fine-tuned using our method. Com$_{\uparrow}$ emphasizes completeness, Con$_{\uparrow}$ emphasizes conciseness, and Bal seeks to balance both. \textit{Succ rate} refers to the proportion of cases where either $S_\text{com}$ or $S_\text{con}$ is non-zero. AvgFaith indicates the average of Faithfulness across different Criteria.}
\label{tb:G-Eval}
\end{table*}


\begin{figure*}[th]
    \centering
    \includegraphics[width=300pt]{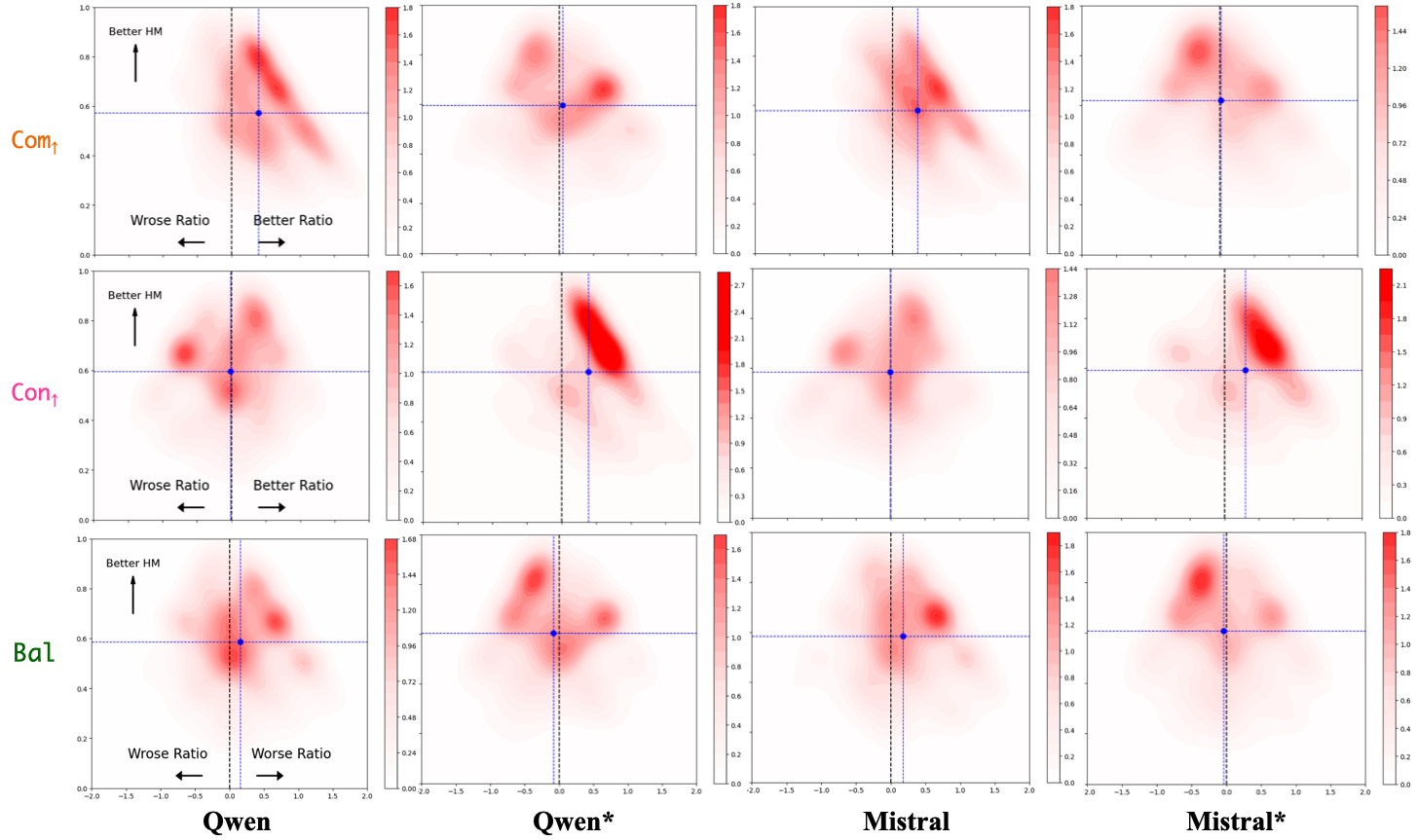}
    \caption{
    Contour plot showing the distribution of models with respect to control ability (x-axis: $\text{R}(\tilde{Y})$) and summary quality (y-axis: $\text{HM}(\tilde{Y})$). Density is represented by color, highlighting regions of strong performance.  
    }
    \label{fig:appendix_1}
\end{figure*}

\begin{figure*}[th]
    \centering
    \includegraphics[width=300pt]{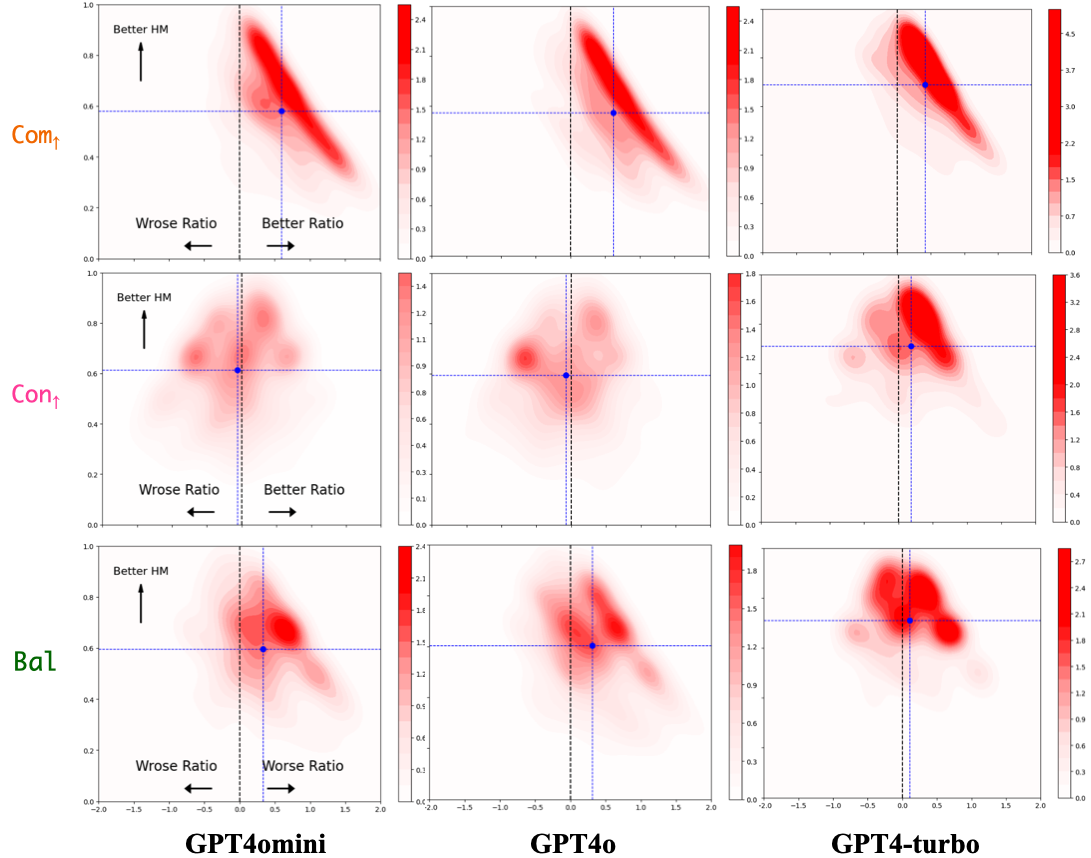}
    \caption{
    Contour plot showing the distribution of models with respect to control ability (x-axis: $\text{R}(\tilde{Y})$) and summary quality (y-axis: $\text{HM}(\tilde{Y})$). Density is represented by color, highlighting regions of strong performance.  
    }
    \label{fig:appendix_contour_openai}
\end{figure*}


\begin{figure*}[th]
    \centering
    \includegraphics[width=400pt]{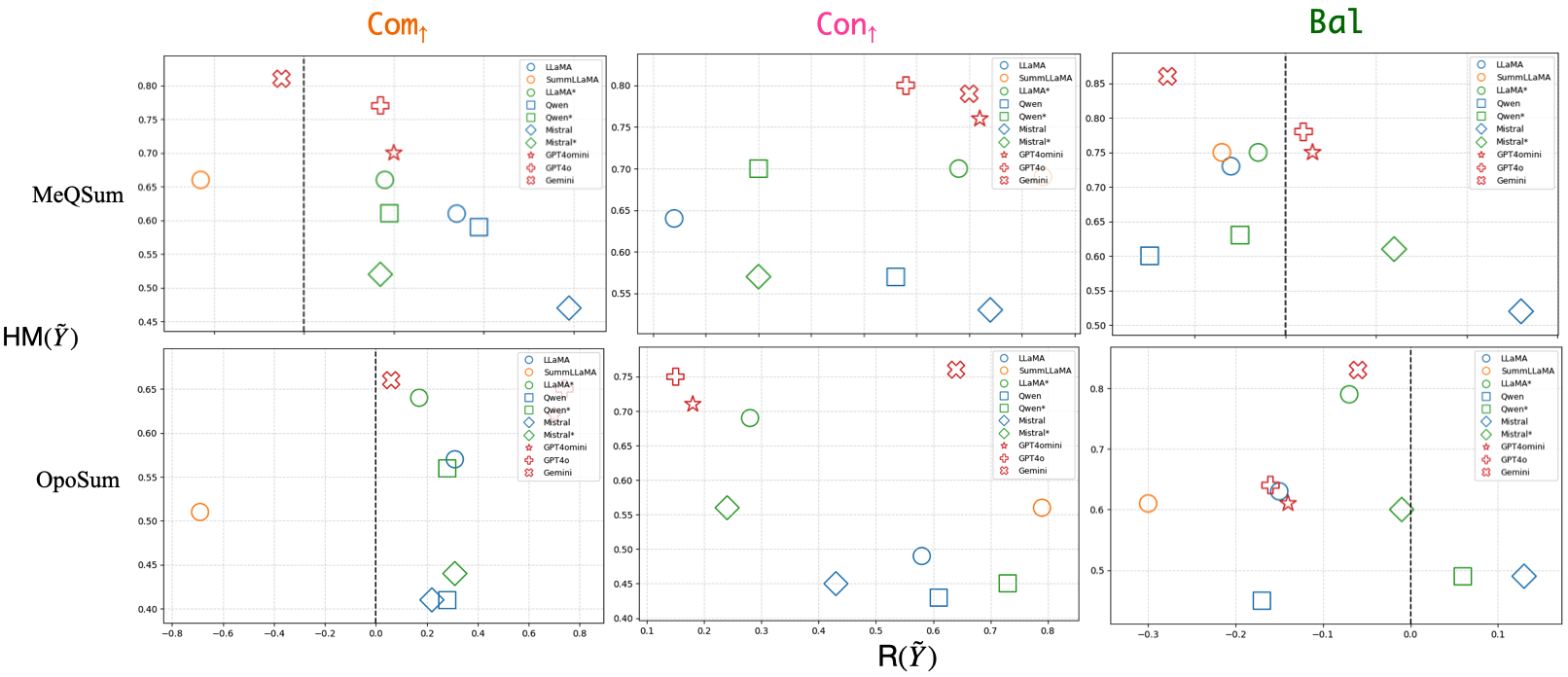}
    \caption{Scatter plot of model performance testing in out-of-domain data (MeQSum; OpoSum) across different control settings. Each point represents the mean of $\text{HM}(\tilde{Y})$ (y-axis) and $\text{R}(\tilde{Y})$ (x-axis) across all test cases in FeedSum test set. Models are grouped by color into four categories: Baseline models (blue), Our methods (green), SummLLaMA (orange), and Commercial models (red). The three panels show performance under different control priorities: Com$_{\uparrow}$ prioritizes completeness, Con$_{\uparrow}$ prioritizes conciseness, and Bal aims to balance both. The vertical dashed line at $\text{R}(\tilde{Y})$ = 0 represents the target equilibrium between completeness and conciseness, providing a reference for model controllability.}
    \label{fig:oo_domain}
\end{figure*}

\begin{table*}[th]
\vspace{1mm}
\centering
\scalebox{0.65}{
\begin{tabular}{lllllllll}
\hline
\textbf{Model}  & \textbf{Source}     & \textbf{Category} & \multicolumn{1}{r}{$\mathbf{S_{com}}$} & \multicolumn{1}{r}{$\mathbf{S_{con}}$} & \multicolumn{1}{r}{$\mathbf{HM}(\tilde{Y})$} & \multicolumn{1}{r}{$\mathbf{R}(\tilde{Y})$} & \multicolumn{1}{r}{\textbf{Succ rate}} & \multicolumn{1}{r}{\textbf{Prop. of 1}} \\ \hline
\multirow{6}{*}{LLaMA}      & \multirow{3}{*}{meqsum} & Com$_{\uparrow}$  & 0.74 $\pm$ 0.18                            & 0.55 $\pm$ 0.23                            & 0.61 $\pm$ 0.18                 & 0.34 $\pm$ 0.54                    & 0.72                                   & 0.24                                    \\
                            &                         & Con$_{\uparrow}$  & 0.66 $\pm$ 0.16                            & 0.7 $\pm$ 0.16                             & 0.64 $\pm$ 0.14                 & 0.07 $\pm$ 0.39                    & 0.78                                   & 0.18                                    \\
                            &                         & Bal               & 0.74 $\pm$ 0.2                             & 0.8 $\pm$ 0.2                              & 0.73 $\pm$ 0.18                 & -0.06 $\pm$ 0.39                   & 0.98                                   & -                                       \\
                            & \multirow{3}{*}{oposum} & Com$_{\uparrow}$  & 0.69 $\pm$ 0.17                            & 0.52 $\pm$ 0.18                            & 0.57 $\pm$ 0.12                 & 0.31 $\pm$ 0.37                    & 0.61                                   & 0.32                                    \\
                            &                         & Con$_{\uparrow}$  & 0.4 $\pm$ 0.16                             & 0.7 $\pm$ 0.17                             & 0.49 $\pm$ 0.13                 & 0.58 $\pm$ 0.41                    & 0.86                                   & 0.11                                    \\
\hline

\end{tabular}
}
\caption{Detailed comparison of model performance using FineSurE. The value shows averages over all cases across two out-of-domain datasets with standard deviations.  $*$ denotes foundation models fine-tuned using our method. Com$_{\uparrow}$ emphasizes completeness, Con$_{\uparrow}$ emphasizes conciseness, and Bal seeks to balance both. \textit{Succ rate} refers to the proportion of cases where either $S_\text{com}$ or $S_\text{con}$ is non-zero. \textit{Proportion of 1} indicates the proportion of cases where both $S_\text{com}$ and $S_\text{con}$ equal 1. In the Com$_{\uparrow}$ and Con$_{\uparrow}$ settings, such cases are considered uncontrollable and are excluded from the computation of $S_\text{com}$, $S_\text{con}$, $\text{HM}(\tilde{Y})$, and $\text{R}(\tilde{Y})$. In contrast, under the Bal setting, these cases are treated as ideal (i.e., perfect) and are retained (denoted as “–”).}
\label{tb:ood}
\end{table*}

\begin{figure*}[th]
    \centering
    \includegraphics[width=400pt]{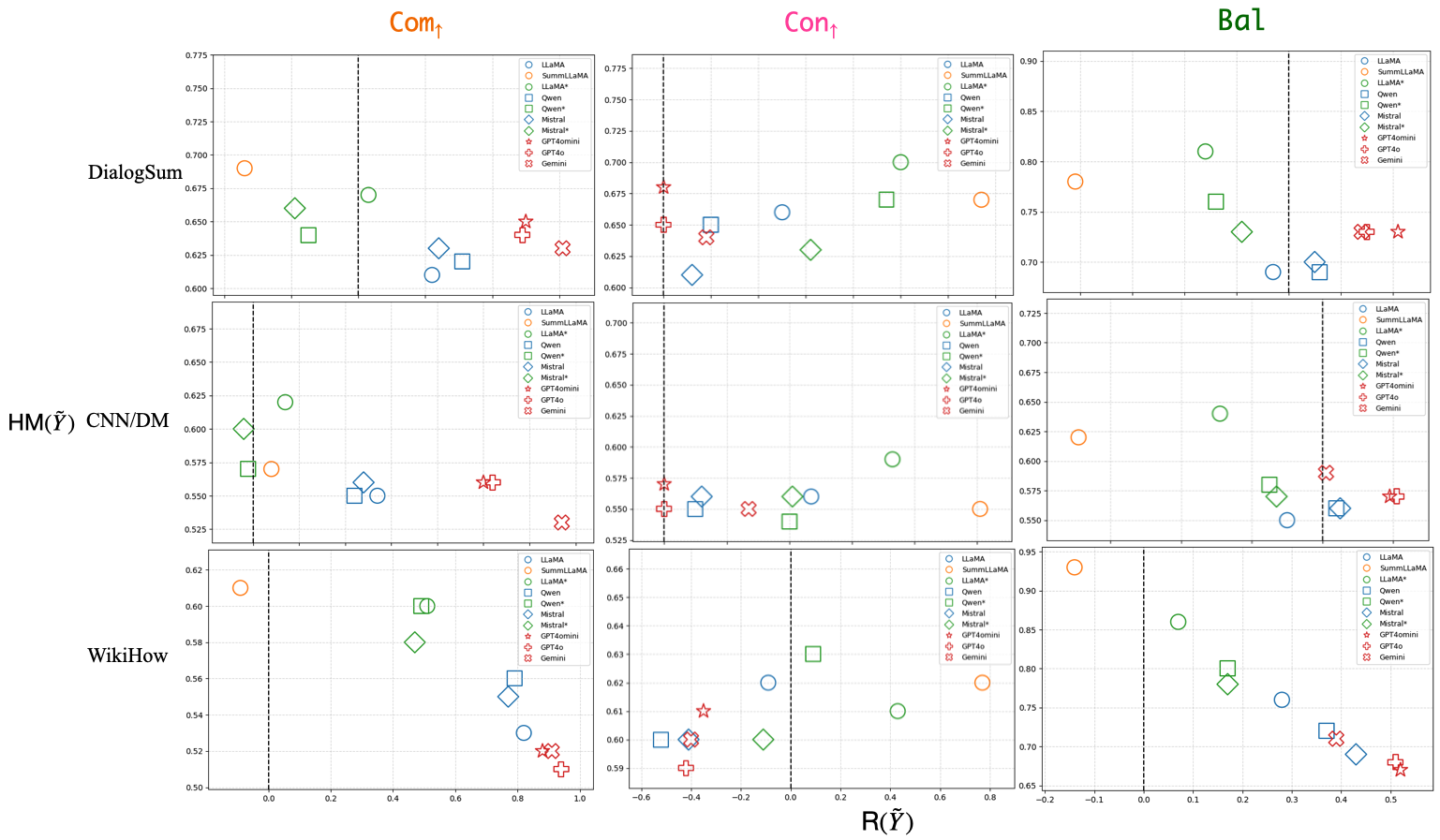}
    \caption{Scatter plot of model performance separated by different domain (DialogSum; CNN/DM; WikiHow) across different control settings. Each point represents the mean of $\text{HM}(\tilde{Y})$ (y-axis) and $\text{R}(\tilde{Y})$ (x-axis) across all test cases in FeedSum test set. Models are grouped by color into four categories: Baseline models (blue), Our methods (green), SummLLaMA (orange), and Commercial models (red). The three panels show performance under different control priorities: Com$_{\uparrow}$ prioritizes completeness, Con$_{\uparrow}$ prioritizes conciseness, and Bal aims to balance both. The vertical dashed line at $\text{R}(\tilde{Y})$ = 0 represents the target equilibrium between completeness and conciseness, providing a reference for model controllability.}
    \label{fig:in_domain}
\end{figure*}

\begin{table*}[th]
\vspace{1mm}
\centering
\scalebox{0.4}{
\begin{tabular}{lllllllll}
\hline
\textbf{Model}  & \textbf{Source}  & \textbf{Criteria} & \multicolumn{1}{r}{$\mathbf{S_{com}}$} & \multicolumn{1}{r}{ $\mathbf{S_{con}}$} & \multicolumn{1}{r}{ $\mathbf{HM}(\tilde{Y})$} & \multicolumn{1}{r}{$\mathbf{R}(\tilde{Y})$} & \multicolumn{1}{r}{\textbf{Succ rate}} & \multicolumn{1}{r}{\textbf{Prop. of 1}} \\ \hline
\multirow{9}{*}{LLaMA}      & \multirow{3}{*}{dialogsum} & Com$_{\uparrow}$  & 0.70 $\pm$ 0.23                            & 0.59 $\pm$ 0.25                            & 0.61 $\pm$ 0.19                 & 0.22 $\pm$ 0.47                    & 0.82                                   & 0.13                                    \\
                            &                            & Con$_{\uparrow}$  & 0.62 $\pm$ 0.19                            & 0.80 $\pm$ 0.25                            & 0.66 $\pm$ 0.16                 & 0.25 $\pm$ 0.5                     & 0.72                                   & 0.22                                    \\
                            &                            & Bal               & 0.71 $\pm$ 0.24                            & 0.74 $\pm$ 0.27                            & 0.69 $\pm$ 0.22                 & -0.03 $\pm$ 0.44                   & 0.94                                   & -                                       \\
                            & \multirow{3}{*}{cnn}       & Com$_{\uparrow}$  & 0.65 $\pm$ 0.21                            & 0.52 $\pm$ 0.22                            & 0.55 $\pm$ 0.19                 & 0.27 $\pm$ 0.48                    & 0.94                                   & 0.00                                    \\
                            &                            & Con$_{\uparrow}$  & 0.48 $\pm$ 0.18                            & 0.76 $\pm$ 0.26                            & 0.56 $\pm$ 0.18                 & 0.47 $\pm$ 0.53                    & 0.94                                   & 0.01                                    \\
\hline
\end{tabular}
}
\caption{Detailed comparison of model performance using FineSurE. The value shows averages over all cases across two in domain datasets with standard deviations. $*$ denotes foundation models fine-tuned using our method. Com$_{\uparrow}$ emphasizes completeness, Con$_{\uparrow}$ emphasizes conciseness, and Bal seeks to balance both. \textit{Succ rate} refers to the proportion of cases where either $S_\text{com}$ or $S_\text{con}$ is non-zero. \textit{Proportion of 1} indicates the proportion of cases where both $S_\text{com}$ and $S_\text{con}$ equal 1. In the Com$_{\uparrow}$ and Con$_{\uparrow}$ settings, such cases are considered uncontrollable and are excluded from the computation of $S_\text{com}$, $S_\text{con}$, $\text{HM}(\tilde{Y})$, and $\text{R}(\tilde{Y})$. In contrast, under the Bal setting, these cases are treated as ideal (i.e., perfect) and are retained (denoted as “–”).}
\label{tb:in_domain}
\end{table*}


\begin{table*}[th]
\vspace{1mm}
\centering
\scalebox{0.8}{
\label{tab:model_comparison}
\begin{tabular}{llcccccc}
\toprule
\textbf{Model} & \textbf{Criteria} & $\mathbf{S_{com}}$        & $\mathbf{S_{fai}}$        & $\mathbf{HM}(\tilde{Y})$   & $\mathbf{R}(\tilde{Y})$    & \textbf{Succ rate}  & \textbf{Prop. of 1} \\
\midrule

\multirow{3}{*}{LLaMA}      & Com$_{\uparrow}$  & 0.63$\pm$0.24 & 0.80$\pm$0.20 & 0.66$\pm$0.18 & -0.30$\pm$0.56 & 0.64 & 0.20 \\
                            & Fai$_{\uparrow}$  & 0.62$\pm$0.22 & 0.80$\pm$0.23 & 0.66$\pm$0.16 & 0.27$\pm$0.50  & 0.63 & 0.24 \\
                            & Bal               & 0.69$\pm$0.27 & 0.82$\pm$0.21 & 0.72$\pm$0.21 & -0.24$\pm$0.45 & 0.87 & -   \\
\cdashline{1-8}
\multirow{3}{*}{SummLLaMA}  & Com$_{\uparrow}$  & 0.63$\pm$0.21 & 0.83$\pm$0.20 & 0.68$\pm$0.16 & -0.31$\pm$0.50 & 0.65 & 0.24 \\
                            & Fai$_{\uparrow}$  & 0.61$\pm$0.18 & 0.84$\pm$0.07 & 0.69$\pm$0.13 & 0.35$\pm$0.35  & 0.62 & 0.26 \\
                            & Bal               & 0.61$\pm$0.31 & 0.86$\pm$0.16 & 0.68$\pm$0.27 & -0.48$\pm$0.63 & 0.89 & -   \\
\cdashline{1-8}
\multirow{3}{*}{LLaMA*}     & Com$_{\uparrow}$  & 0.77$\pm$0.23 & 0.73$\pm$0.19 & 0.73$\pm$0.16 & 0.04$\pm$0.39  & 0.79 & 0.09 \\
                            & Fai$_{\uparrow}$  & 0.72$\pm$0.24 & 0.86$\pm$0.15 & 0.76$\pm$0.16 & 0.22$\pm$0.38  & 0.70 & 0.18 \\
                            & Bal              & 0.75$\pm$0.25 & 0.81$\pm$0.18 & 0.75$\pm$0.19 & -0.12$\pm$0.40 & 0.89 & -   \\
\hline

\end{tabular}}
\caption{Detailed comparison of model performance using FineSurE. The value shows averages over all cases in FeedSum with standard deviations. $*$ indicates the foundation model fine-tuned using our method. Com$_{\uparrow}$ prioritizes faithfulness, Fai$_{\uparrow}$ prioritizes faithfulness, and Bal aims to balance both. \textit{Succ rate} denotes the proportion of cases where either $S_\text{com}$ or $S_\text{fai}$ is non-zero. \textit{Proportion of 1} indicates the proportion of cases where both $S_\text{com}$ and $S_\text{fai}$ equal 1. In the Com$_{\uparrow}$ and Fai$_{\uparrow}$ settings, these are considered uncontrollable cases and are excluded when computing $S_\text{com}$, $S_\text{con}$, $\text{HM}(\tilde{Y})$, and $\text{R}(\tilde{Y})$. In contrast, under Bal, these cases are treated as ideal ({\em i.e.}, perfect), so they are retained (denoted as “–”).}
\label{tab:model_com_fai}
\end{table*}

\begin{figure*}[t]
    \centering
    \includegraphics[width=400pt]{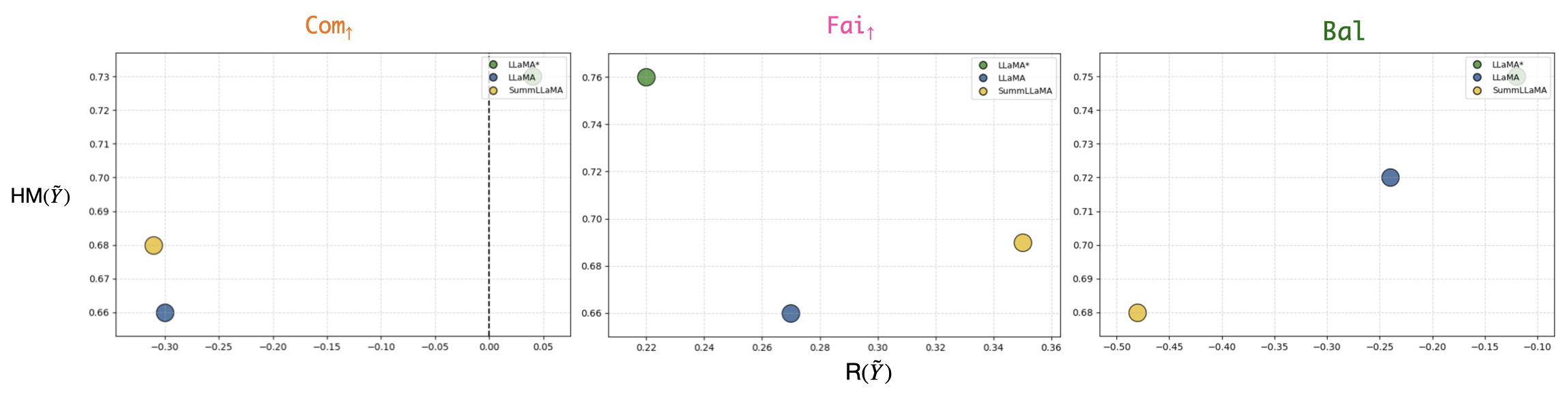}
    \vspace{-5pt}
    \caption{\small Model performance across different control settings. Each point represents the mean of $\text{HM}(\tilde{Y})$ (y-axis) and  $\text{R}(\tilde{Y})$ (x-axis) across all test cases in FeedSum test set. The three panels show performance under different control priorities: \textcolor[HTML]{ff7f0e}{Com$_\uparrow$} prioritizes completeness, \textcolor[HTML]{E22B98}{Fai$_\uparrow$}  prioritizes faithfulness, and \textcolor[HTML]{009901}{Bal} aims to balance both. The vertical dashed line at  $\text{R}(\tilde{Y})$ = 0 represents the controllability target (reference) between completeness and conciseness.}
    \label{fig:com_fai_comparison}
    \vspace{-4mm}
\end{figure*}

\begin{figure*}[t]
    \centering
    \includegraphics[width=0.47\textwidth]{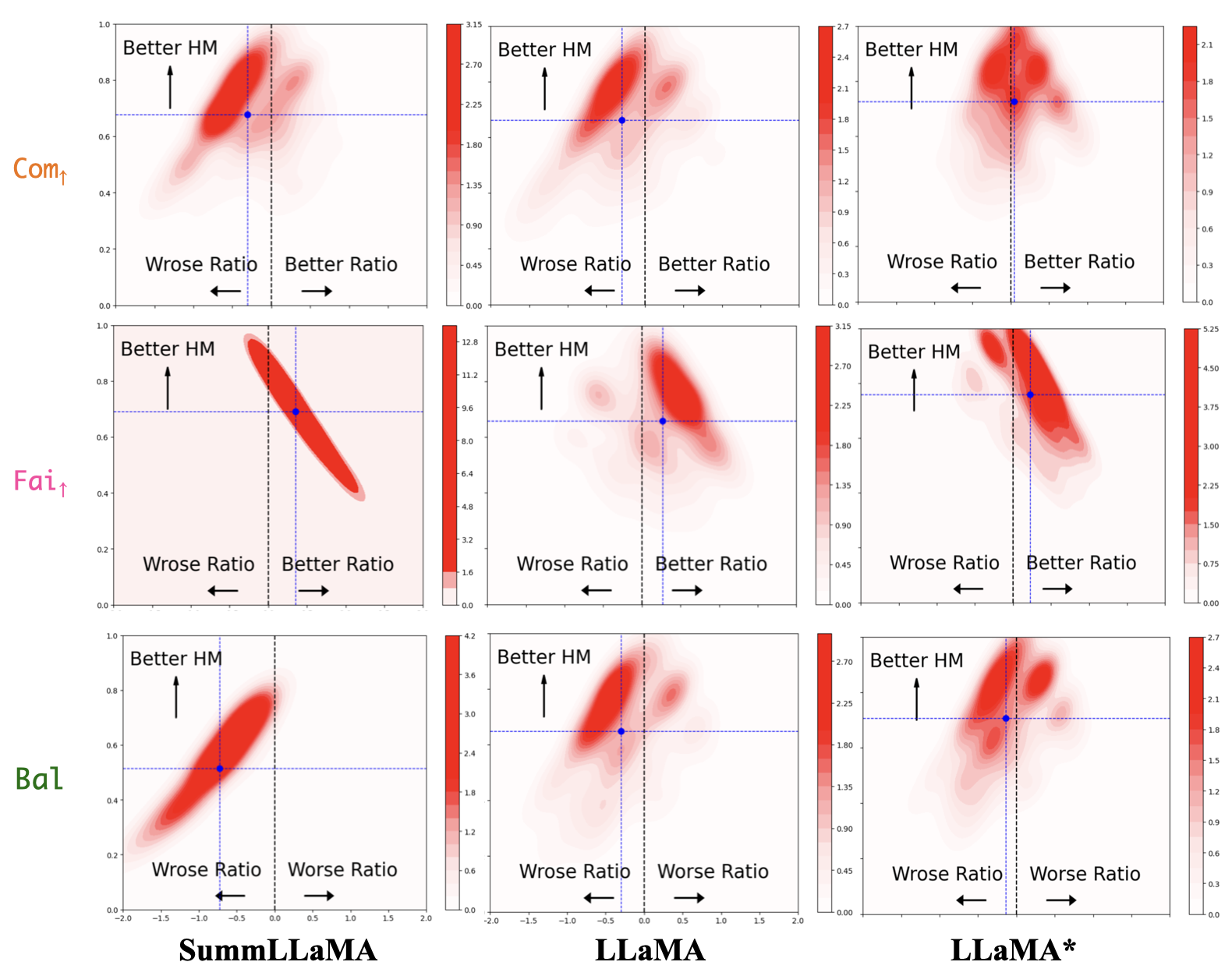}
    \vspace{-2mm}
    \caption{\small Distributions of  $\text{R}(\tilde{Y})$ (x) and $\text{HM}(\tilde{Y})$ (y) metrics.
    \textcolor[HTML]{ff7f0e}{Com$_\uparrow$} prioritizes completeness, \textcolor[HTML]{E22B98}{Fai$_\uparrow$}  prioritizes faithfulness, and \textcolor[HTML]{009901}{Bal} to balances them.
    \textcolor[HTML]{0000FF}{Blue dashed lines}
     mark the mean of the metrics, and the arrows point in the direction in which metrics are better or worse.}
    \label{fig:contour_com_fai}
    \vspace{-6mm}
\end{figure*}

\end{document}